\documentclass{article}

\usepackage{PRIMEarxiv}

\usepackage[utf8]{inputenc} 
\usepackage[T1]{fontenc}    
\usepackage{hyperref}       
\usepackage{url}            
\usepackage{booktabs}       
\usepackage{amsfonts}       
\usepackage{nicefrac}       
\usepackage{microtype}      
\usepackage{lipsum}
\usepackage{fancyhdr}       
\usepackage{graphicx}       
\graphicspath{{media/}}     

\usepackage[numbers]{natbib}

\usepackage[nolist,nohyperlinks]{acronym}
\usepackage{siunitx}
\usepackage{subcaption}
\usepackage{threeparttable}
\usepackage{enumitem}
\usepackage{amsmath}
\usepackage{multirow}

\title{Latent PDE mapping for efficient physics-informed learning across geometries with limited data}

\author{
Ingvild Askim Adde$^{1}$,
Mary M. Maleckar$^{2}$,
Gabriel Balaban$^{1,2}$\\[1ex]
$^{1}$Faculty of Health and Technology, Kristiania University of Applied Sciences, Oslo, Norway\\
$^{2}$Department of Computational Physiology, Simula Research Laboratory, Oslo, Norway\\
}

\begin{document}
\maketitle

\begin{abstract}
In this study, we introduce latent PDE mapping, a broadly applicable physics-informed learning technique designed to enable efficient geometric generalization with sparse training data. Latent PDE mapping pulls back geometry-specific PDE residuals and boundary conditions to a predefined latent geometry via the deformation gradient, thereby enabling the automated calculation of geometry-consistent shape gradients that are missing in conventional physics-informed machine learning formulations. We demonstrate the utility of latent PDE mapping in solving the anisotropic Aliev-Panfilov PDE of cardiac electrophysiology using both physics-informed neural networks and physics-informed deep operator networks. The Aliev-Panfilov PDE serves as a challenging exemplar: a nonlinear, time-dependent PDE benchmark with sharp gradients that are expensive to capture using traditional numerical solvers. To represent the limited data regime, we train the networks using just fifteen geometric samples drawn from parameterized distributions in two and three spatial dimensions. While modest improvements appear for geometries parameterized by affine and shear deformations, latent PDE mapping demonstrates significant benefits on select geometric families, achieving a factor \textasciitilde{}4-6 reduction in mean relative $L_2$ error. Furthermore, our results show that the computational cost of applying latent PDE mapping was modest during network training, and negligible at inference. Taken together, our study highlights how latent PDE mapping facilitates the creation of generalizable physics-informed machine learning models from limited sets of training geometries.
\end{abstract}


\begin{acronym}
    \acro{FEM}{finite element method}
    \acro{PINN}{physics-informed neural network}
    \acro{PA-PINN}{parameter aware physics-informed neural network}
    \acro{PI-DON}{physics-informed deep operator network}
    \acro{PDE}{partial differential equation}
    \acro{RMSE}{root mean squared error}
    \acro{AT}{activation time}
    \acro{RT}{repolarization time}
    \acro{CVD}{cardiovascular disease}
    \acro{SCD}{sudden cardiac death}
    \acro{FDM}{finite difference method}
    \acro{FVM}{finite volume method}
    \acro{NN}{neural network}
    \acro{ML}{machine learning}
    \acro{FCNN}{fully connected neural network}
    \acro{DNN}{deep neural network}
    \acro{EP}{electrophysiology}
    \acro{BC}{boundary condition}
    \acro{IC}{initial condition}
    \acro{PI-CNN}{physics-informed convolutional neural network}
    \acro{PI-GNN}{physics-informed graph neural network}
    \acro{PCA}{principal component analysis}
    \acro{SDF}{signed distance function}
    \acro{MSE}{mean squared error}
    \acro{PCA}{principal component analysis}
    \acro{LPM}{Latent PDE Mapping}
    \acro{BC}{boundary condition}
\end{acronym}

\newcommand{\vx}{\boldsymbol{x}}
\newcommand{\vn}{\boldsymbol{n}}
\newcommand{\vX}{\boldsymbol{X}}
\newcommand{\vU}{\boldsymbol{U}}
\newcommand{\vN}{\boldsymbol{N}}
\newcommand{\tF}{\mathbf{F}}
\newcommand{\tI}{\mathbf{I}}
\newcommand{\tD}{\mathbf{D}}
\newcommand{\vA}{\boldsymbol{A}}
\newcommand{\vM}{\boldsymbol{M}}

\newcommand{\Gexp}{$\mathcal{G}_{exp}$}
\newcommand{\Gshear}{$\mathcal{G}_{shear}$}
\newcommand{\Gnonlin}{$\mathcal{G}_{nonlin}$}
\newcommand{\Grot}{$\mathcal{G}_{rot}$}

\newcommand{\GexpExt}{$\mathcal{G}_{exp}^*$}
\newcommand{\GshearExt}{$\mathcal{G}_{shear}^*$}
\newcommand{\GnonlinExt}{$\mathcal{G}_{nonlin}^*$}
\newcommand{\GrotExt}{$\mathcal{G}_{rot}^*$}

\newcommand{\Hrotx}{$\mathcal{H}_{rot}^{x}$}
\newcommand{\Hroty}{$\mathcal{H}_{rot}^{y}$}
\newcommand{\Hrotz}{$\mathcal{H}_{rot}^{z}$}

\newcommand{\HrotxExt}{$\mathcal{H}_{rot}^{x*}$}
\newcommand{\HrotyExt}{$\mathcal{H}_{rot}^{y*}$}
\newcommand{\HrotzExt}{$\mathcal{H}_{rot}^{z*}$}

\newcommand{\NN}{NN}
\newcommand{\PINN}{Basic-PINN}
\newcommand{\Affine}{PA-PINN}
\newcommand{\LG}{LG-PINN}
\newcommand{\LPM}{LPM-PINN}

\newcommand{\DeepO}{Basic-DON}
\newcommand{\LGDeepO}{LG-DON}
\newcommand{\LPMDeepO}{LPM-DON}

\newcommand{\refgeom}{latent geometry}

\section{Introduction}
Physics-informed machine learning has emerged as a powerful paradigm for solving complex physical problems across a wide range of engineering tasks, including fluid dynamics \cite{ali2025machine, galazis2025pinning}, biomechanics \cite{sotiropoulos2026deeponet, arzani2021uncovering}, and solid mechanics \cite{wang2023exact, abueidda2021meshless}. A common strategy for developing physics-informed models is to incorporate governing physical laws through soft constraints. In this setting, governing equations are embedded by augmenting data-driven loss functions with \ac{PDE} residuals and \ac{BC} terms, providing a flexible mechanism for enforcing physical consistency during learning \cite{raissi2019physics}. Once trained, these physics-informed models can rapidly solve parameterized \acp{PDE}, providing data-informed simulations of dynamical systems orders of magnitude faster than traditional numerical solvers. This is a crucial property in several applied engineering problems, where fast and accurate updates over time are necessary. Crucially, engineering challenges across several domains may require updates across multiple, diverse geometries. For instance, structural optimization of an airfoil demands several simulations across various shapes to design the optimal drag-to-lift ratio \cite{lyu2015aerodynamic}, and patient-specific treatment plans within medical digital twins require numerous, swift updates as the underlying physiological geometry changes \cite{niederer2021scaling}. However, conventional formulations of physics-informed models \cite{raissi2019physics, goswami2023physics} require costly retraining when encountering novel geometries, thereby undermining the computational benefits that motivate their use. 

A common approach for handling variable geometries when applying physics-informed learning has been to develop specialized network architectures, replacing multilayer perceptrons with physics-informed convolutional neural networks \citep{gao2021phygeonet}, physics-informed graph neural networks  \citep{dalton2023physics, peng2023physics, wurth2024physics, gao2022physics}, or physics-informed PointNet \citep{kashefi2022physics}. These architectures are better suited to handling variable geometries than basic fully-connected neural networks, but require uniform grids, demand complex meshing at inference, or struggle to generalize across \ac{PDE} parameters \citep{zhong2025physics}. Consequently, their applicability remains constrained to scenarios involving complex geometries requiring non-uniform discretizations or applications needing fast inference while maintaining robustness to unseen \ac{PDE} parameters. To overcome these challenges, physics-informed frameworks have been augmented with shape descriptors \citep{regazzoni2022universal, costabal2024delta} or global geometric parameters \citep{sun2023physics, zhong2025physics}. In this way, information about the geometric dependency is provided as an additional input to neural networks. While showing promise, these methods still formulate their physics losses directly on the varying physical geometries, which limits the gradient information available to networks during training.

Another approach involves combining physics-based losses with latent geometries, where inputs are embedded into a common latent space to facilitate comparison and efficient representation learning across different shapes. Autoencoder-based approaches have been increasingly used as a means for encoding geometric variability and \ac{PDE} parameters into low-dimensional latent spaces \cite{taufik2025latentpinns, karumuri2026physics, oldenburg2022geometry}. Such approaches provide flexible and trainable representations that facilitate learning across geometries. Nevertheless, they introduce additional pre-training stages and hyperparameters, further complicating an already challenging optimization problem. Moreover, validating whether the learned latent space faithfully represents complex geometries remains non-trivial, thereby limiting models' interpretability. As noted by \citet{tran2025artificial}, usability, explainability, and transparency are key factors shaping clinicians' trust and engagement with machine learning models, highlighting the need for simpler modeling approaches that are grounded in mathematical formulations.

A complementary line of work addresses geometric variability by introducing predefined latent geometries in a more structured and transparent manner than autoencoder-based approaches. \citet{regazzoni2022universal} proposed a universal latent space for parameterized geometries, enabling learning across varying shapes by using a latent geometry. Similarly, \citet{mezzadri2023framework} introduced a framework that aligns geometric variability through latent embeddings, enabling simple linear elasticity models to generalize across freeform domains. More recently, \citet{burbulla2023physics} introduced a \ac{PDE} mapping to low-dimensional manifolds and applied it to simple linear \acp{PDE}. However, most current physics-informed latent-geometry methods \cite{mezzadri2023framework, regazzoni2022universal, burbulla2023physics} are either limited to simple, linear, static \acp{PDE} or do not map the \ac{PDE} itself. This limits the utility of the methods in real-world applications, which are often complex, nonlinear, and dynamic. While a recent study by \citet{JacobiNet} has introduced a neural network mapping approach to handling flexible geometries, the use of the mapping network makes the method data hungry, requiring a large number of geometrical examples to learn an accurate geometry map. To the best of our knowledge, no work has yet shown that physics-informed learning with mapped \acp{PDE} can generalize well to geometries outside of the training distribution using limited geometrical examples.

In our study, we introduce \textit{latent PDE mapping}, a mapping framework that pulls back geometry-specific \ac{PDE} residuals and \ac{BC} terms to a preselected latent geometry, thereby facilitating accurate shape-gradient calculations. Latent \ac{PDE} mapping builds on recent physics-informed machine learning studies involving latent geometries \cite{li2023fourier, li2023geometry, yin2024scalable, zhong2025physics, mezzadri2023framework, regazzoni2022universal, burbulla2023physics}, while employing principles from nonlinear solid mechanics \cite{HolzaphelSolidMechanicsBook} commonly used in the \ac{FEM} literature \cite{nobile2012active}. Unlike the JacobiNet \cite{JacobiNet}, our mappings are predetermined so that no additional networks need to be trained. In this way, our framework allows for efficient and automated calculation of geometry parameter gradients during back-propagation, which are typically missing in standard physics-informed formulations. This advancement leads to a broadly applicable mathematical framework that moves beyond simple \ac{PDE} mapping to physics-informed formulations with geometrically variable shapes and nonlinear, time-dependent \acp{PDE} in the limited data regime. 

Our key contributions can be summarized as follows:

\begin{itemize}
    \item We introduce latent \ac{PDE} mapping, a novel technique that maps geometry-specific \ac{PDE} and \ac{BC} residuals to a shared latent geometry, enabling physics-informed frameworks with soft constraints to learn meaningful representations across diverse geometrical shapes using minimal training data.
    \item We implement latent PDE mapping in two representative physics-informed learning frameworks, namely \acp{PINN} and \acp{PI-DON}, using the Aliev–Panfilov model of cardiac electrophysiology, a challenging nonlinear, time-dependent \ac{PDE} with sharp gradients and anisotropic conditions.
    \item We provide theoretical and empirical evidence that latent \ac{PDE} mapping properly accounts for geometric variability in the physics loss gradient, yielding more generalizable representations for neural networks trained with physics-informed losses.
\end{itemize}

In this work, we test latent \ac{PDE} mapping using \ac{PINN} \cite{raissi2019physics} and \ac{PI-DON} \cite{wang2021learning} architectures. However, our framework is an architecture-agnostic formulation that can be integrated with a broad range of neural network backbones, including graph neural networks \cite{dalton2023physics, peng2023physics, wurth2024physics, gao2022physics}, neural operator networks \cite{li2023fourier, li2023geometry, yin2024scalable, zhong2025physics}, and fully-connected networks \cite{raissi2019physics, sun2023physics, regazzoni2022universal}. Due to its incorporation of the deformation gradient, our approach naturally extends to nonlinear physics problems defined over diffeomorphic geometries. Such problems appear in a wide range of applications, from biomedical transport to airplane design.

The remaining parts of the paper are organized as follows: Section~\ref{sec:methods} introduces the proposed methodology, including the latent PDE mapping framework with a mathematical explanation of how it improves geometric generalizability during training (Section \ref{sec:latent-pde-mapping}), 2D and 3D geometries (Section \ref{sec:geometries}), the synthetic electrophysiology data (Section \ref{sec:ep_data}), the neural network architectures (Section \ref{sec:neural_networks}), and the training procedures (Section \ref{sec:training_procedures}). The experimental results are presented in Section~\ref{sec:experiments}, followed by a discussion of the findings, limitations, and directions for future work in Section~\ref{sec:discussion}. Finally, Section~\ref{sec:conclusion} concludes the paper.

\section{Methods}
\label{sec:methods}

\begin{figure}
    \centering
    \includegraphics[width=0.9\linewidth]{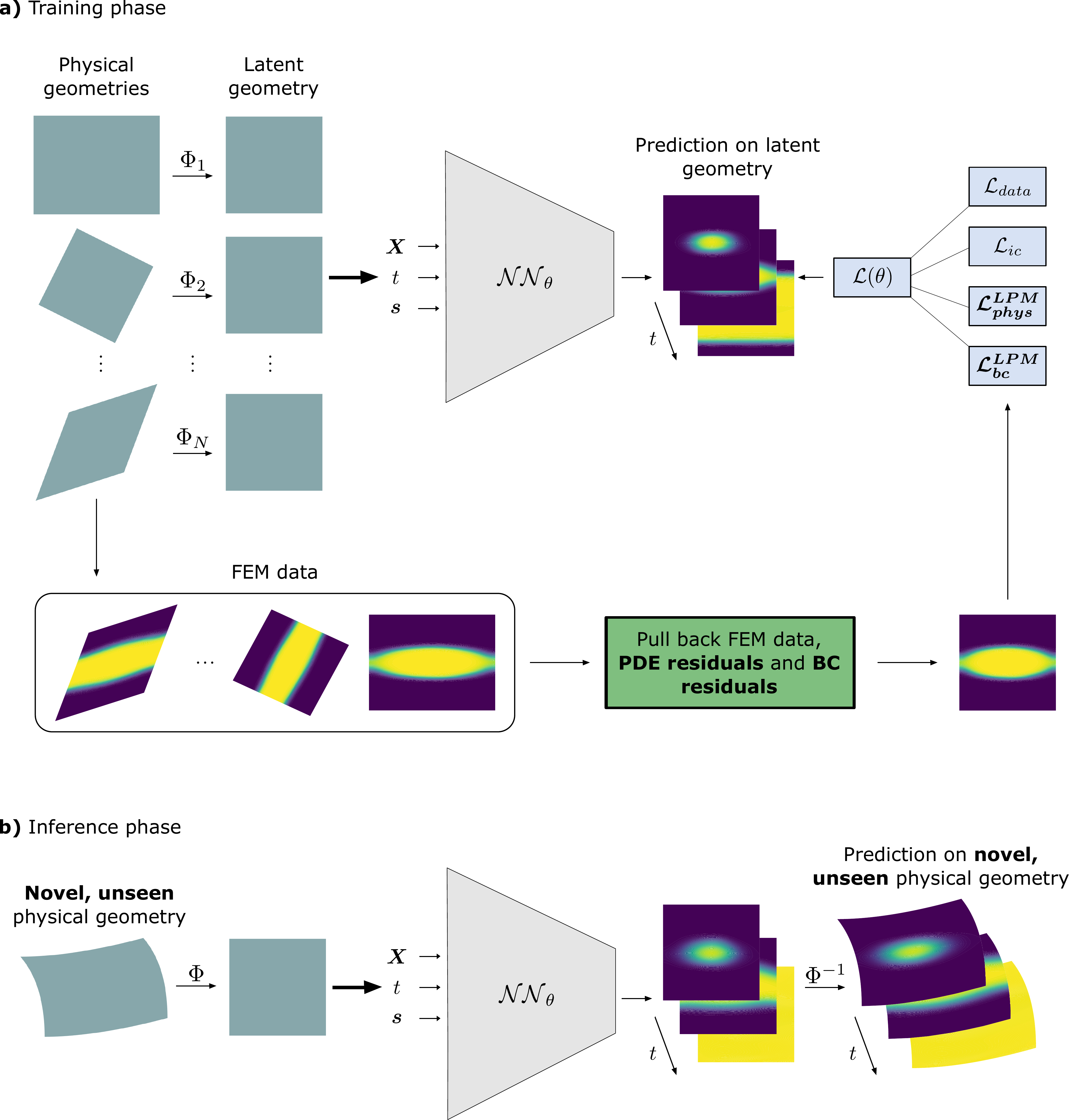}
    \caption{
    Latent \ac{PDE} mapping (LPM) enables accurate predictions on novel, unseen geometries. (a) Training phase: Spatiotemporal data from each physical geometry are mapped to a predefined latent geometry via a diffeomorphic mapping ($\Phi_N$), and passed to a neural network ($\mathcal{NN}_{\theta}$) alongside shape parameters ($\boldsymbol{s}$) encoding each geometry. FEM data, \ac{PDE} residuals, and BC residuals defined on the physical geometries are pulled back to the latent geometry through the deformation gradient and the Jacobian, yielding a consistent latent representation. This shared representation ensures accurate shape gradients when minimizing the physics and boundary condition loss functions ($\mathcal{L}_{phys}^{LPM}, \mathcal{L}_{bc}^{LPM}$) during training. (b) Inference phase: Network predictions are mapped back to the physical geometry via the inverse mapping ($\Phi^{-1}$), enabling generalization to geometries outside the training distribution.}
    \label{fig:overview}
\end{figure}

\subsection{Latent PDE mapping}
\label{sec:latent-pde-mapping}
Figure~\ref{fig:overview} presents an overview of our proposed latent \ac{PDE} mapping technique. Mathematically, we consider a time-dependent \ac{PDE} defined over a geometry $\Omega(s)$. Here, $s$ is a set of shape parameters describing the overall geometry of $\Omega$. The governing \ac{PDE} is given as
\begin{equation}
    \mathcal{F}\left( u\left(\vx,t \right) \right) = f(\vx, t, u), \hspace{0.4cm} (\vx, t) \in \Omega(s) \times \mathcal{T}
    \label{eq:generic-pde}
\end{equation}
where $\mathcal{F}$ denotes a differential operator,  $f$ represents a source term that introduces external influences into the system, $\vx \in \Omega(s) \subset \mathbb{R}^d$ are the spatial coordinates, $t \in \mathcal{T} \subset \mathbb{R}$ is the time, and $u$ is the unknown \ac{PDE} solution. In practice, obtaining an exact solution to \eqref{eq:generic-pde} is often intractable due to the complexity of the underlying system. To address this, we employ a neural network $\mathcal{NN}$ to approximate the solution such that
\begin{equation}
    \mathcal{NN}(\vx, t, s; \theta) = u_{\theta} \approx u(\vx,t; s)
\end{equation}
where $\theta$ represents the trainable network parameters. Furthermore, we enforce physics-informed learning through soft constraints by embedding \ac{PDE} residuals in the loss function. The residual is defined as
\begin{equation}
    \mathcal{R}(\vx, t, u) = \mathcal{F}\left( u\left(\vx,t \right) \right) - f(\vx, t, u) = 0, \hspace{0.25cm} (\vx, t) \in \Omega(s) \times \mathcal{T}
    \label{eq:residual}
\end{equation}
where $\mathcal{R}$ depends on the shape parameters $s$ implicitly via the physical geometry $\Omega(s)$. With latent \ac{PDE} mapping, we draw inspiration from methods within the \ac{FEM} and solid continuum mechanics literature \citep{HolzaphelSolidMechanicsBook, nobile2012active}, and express the geometry-specific residual in \eqref{eq:residual} over a predefined latent geometry. Thus, we assume that a continuous map exists between $\Omega(s)$ and the selected latent geometry $\Omega_0$, defined as
\begin{equation}
    \Phi(s) := \vx(s) \rightarrow \vX
    \label{eq:mapping}
\end{equation}
where $\vx(s)$ is a given point in $\Omega(s)$ while $\vX$ is the associated point in $\Omega_0$. Physical quantities can be mapped from $\Omega(s)$ to $\Omega_0$, or vice versa, through the deformation gradient and deformation Jacobian, given in their most general form as
\begin{equation}
    \tF(\vX, s) = \nabla_{\vX} \Phi^{-1} (\vX, s)
    \label{eq:F}
\end{equation}
and $ J(\vX, s) = \text{det}(\tF)$, respectively \citep{HolzaphelSolidMechanicsBook}. In this study, we use the deformation gradient to map the geometry-specific $\mathcal{R}$ in \eqref{eq:residual} to a shared latent representation, yielding
\begin{equation}
    \mathcal{R}^{LPM}(\vX, t, u, \tF, J;s) = \mathcal{F}\left( u, \tF, J; s \right) 
    - f(\vX, t, u, \tF, J; s), \hspace{0.4cm} (\vX, t) \in \Omega_0 \times \mathcal{T}.
    \label{eq:generic-pde-mapped}
\end{equation}
In this way, the residual dependence on $s$ has been moved from the physical geometry $\Omega(s)$ into the \ac{PDE} itself through the deformation gradient $\tF$ and deformation Jacobian $J$. This approach is what we refer to as the \emph{latent \ac{PDE} mapping} technique.

\subsubsection{Application to nonlinear, time-dependent, stiff systems: the Aliev-Panfilov PDE}
We demonstrate the latent \ac{PDE} mapping technique on the Aliev-Panfilov model from cardiac electrophysiology. The Aliev-Panfilov \ac{PDE} \citep{aliev1996simple} is used to describe the evolution of the transmembrane potential $V$ over a physical geometry representing cardiac tissue. It represents a challenging \ac{PDE} system well due to its nonlinearity, sharp gradients, and time-dependency. In mathematics, the Aliev-Panfilov \ac{PDE} is
\begin{equation}
    \begin{cases}
        \frac{\partial V}{\partial \tau} = \nabla \cdot (\tD \nabla V) - kV(V - a)(V - 1) - VW & \text{in } \Omega(s),  \\
        \frac{\partial W}{\partial \tau} = \left( \epsilon_0 + \frac{\mu_1 W}{V + \mu_2} \right) \left( -W-kV \left( V-a-1 \right) \right) & \text{in } \Omega(s), \\
        \tD  \nabla V \cdot \vn = 0 & \text{on } \partial \Omega(s). \\
    \end{cases}
    \label{eq:ap}
\end{equation}
Here, $\Omega(s)$ is a geometry parameterized by $s$, $\vn$ is the boundary normal, and $V, W,$ and $\tau$ are dimensionless variables representing the transmembrane potential, recovery variable, and time, respectively. $V\in [0, 1]$ is given in arbitrary units (AU), while $\tau = 12.9 t$ is measured in temporal units (TU) with $t$ given in milliseconds. The tissue conductivity is defined by the diffusion tensor $\tD$, while $k, a, \epsilon_0, \mu_1, \mu_2$ are parameters controlling the overall shape and temporal dynamics of $V$ and $W$. Also, we employ a no-flux Neumann boundary condition to ensure that there is no leakage of $V$ to regions outside of $\Omega(s)$.

By applying latent \ac{PDE} mapping, the Aliev-Panfilov \ac{PDE} in \eqref{eq:ap} is mapped to a latent representation in $\Omega_0$
\begin{equation}
    \begin{cases}
        \frac{\partial V}{\partial \tau} = \frac{1}{J} \nabla \cdot (J \tF^{-1} \tD \tF^{-T} \nabla V) - kV(V - a)(V - 1) - VW & \text{in } \Omega_0,  \\
        \frac{\partial W}{\partial \tau} = \left( \epsilon_0 + \frac{\mu_1 W}{V + \mu_2} \right) \left( -W-kV \left( V-a-1 \right) \right) & \text{in } \Omega_0, \\
        J \tF^{-1} \tD \tF^{-T} \nabla V \cdot \vN = 0 & \text{on } \partial \Omega_0, \\
    \end{cases}
    \label{eq:mapped_ap}
\end{equation}
where $\vN$ is the boundary normal vector of $\Omega_0$. Note that the geometry dependence on $s$ has been moved to the \ac{PDE} itself via the deformation gradient $\tF = \tF(\vX, s)$ and the Jacobian $J = J(\vX, s)$. This enables accurate shape-gradient calculations during physics-informed back-propagation. A detailed derivation of \eqref{eq:mapped_ap} can be found in Appendix~\ref{appendix:mapped_ap}.

\subsubsection{Accurate gradient calculation with latent PDE mapping}
\label{subsec:shape-gradients}
The physics loss during training is typically evaluated with the \ac{MSE} of $\mathcal{R}$ \citep{wang2023expert}, given as
\begin{equation}
    \mathcal{L}_{phys} = \frac{1}{N_{phys}} \sum_i^{N_{phys}} \mathcal{R}_i^2
    \label{eq:mse_R}
\end{equation}
using a traditional mini-batch approach with $N_{phys}$ collocation points to evaluate $\mathcal{R}$. This approach treats $\mathcal{R}$ as independent of $\Omega(s)$ during optimization, which is not the case and can lead to inaccurate gradient estimates. Thus, a more accurate formulation is to evaluate the physics loss as a continuous integral
\begin{equation}
    \mathcal{L}_{phys} = \int_{\Omega(s)} \mathcal{R}(\vx, t, u)^2 \text{d}\Omega
    \label{eq:continuous-mse}
\end{equation}
and to apply the Leibniz integral rule when computing the shape gradient $\partial \mathcal{L}_{phys}/{\partial s}$. This results in
\begin{equation}
    \frac{\partial \mathcal{L}_{phys}}{\partial s} = \int_{\Omega(s)} \frac{\partial}{\partial s}  \mathcal{R}(\vx, t, u)^2 \text{d}\Omega + \int_{\partial\Omega(s)} \mathcal{R}(\vx, t, u)^2 \frac{\partial \vx}{\partial s} \cdot \vn \text{d}S
    \label{eq:reynold}
\end{equation}
where $\vn$ is the outward unit normal to the boundary $\partial \Omega(s)$ and $\text{d}S$ is an infinitesimally small part of the boundary. The second term in \eqref{eq:reynold} accounts for the movement of the boundary, which is neglected in the discrete loss formulation in \eqref{eq:mse_R}.

With latent \ac{PDE} mapping, the dependency of $s$ is moved from the geometry into the \ac{PDE} itself via the deformation gradient $\tF = \tF(\vX, s)$.  Consequently, the integrand does not vary with $s$ and the shape gradient can be computed directly
\begin{equation}
    \frac{\partial \mathcal{L}^{LPM}_{phys}}{\partial s} = \int_{\Omega_0} \frac{\partial}{\partial s}  \mathcal{R}^{LPM}(\vX, t, u, \tF, J; s)^2 \text{d}\Omega_0.
    \label{eq:mapped-reynold}
\end{equation}
Thus, the straightforward \ac{MSE} in \eqref{eq:mse_R} can be applied during training without sacrificing gradient accuracy. Based on these considerations, we hypothesized that improving the accuracy of the physics loss gradient via latent PDE mapping can improve the generalizability of physics-informed machine learning models to novel geometries.

\subsection{Parametrized families of geometries}
\label{sec:geometries}
\begin{figure}
  \begin{center}
    \centerline{\includegraphics[width=0.8\linewidth]{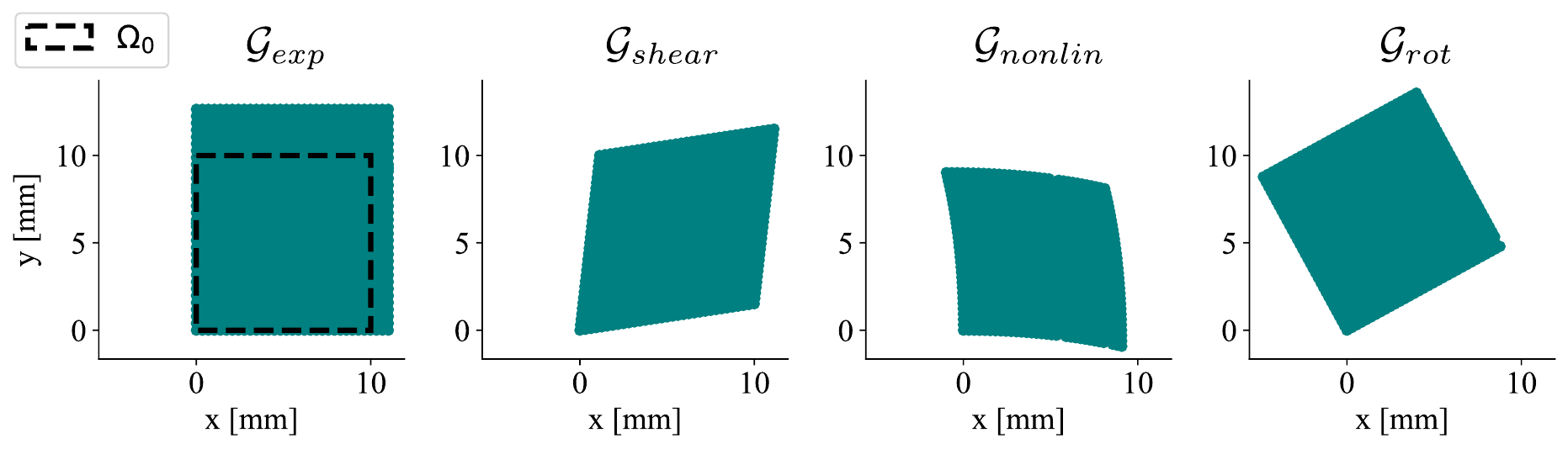}}
    \caption{
      The figure shows an example geometry from \Gexp, \Gshear, \Gnonlin, and \Grot. The dashed line illustrates the latent geometry $\Omega_0$.
    }
    \label{fig:geometries-example}
  \end{center}
\end{figure}

We constructed parameterized families of geometries in 2D and 3D to train and test our physics-informed models. These parameterized families were derived from two latent geometries that we chose beforehand: a simple $10 \times 10$ mm square in 2D, and a $10 \times 10 \times 10$ mm cube in 3D. We created families of geometries by applying various deformations to these latent geometries. In 2D, these deformations had a unified general form
\begin{equation}
    \vx(s) = \vA \vX + \vX^T \vM \vX,
    \label{eq:affine_transform}
\end{equation}
with
\begin{equation}
    \vA = \begin{bmatrix}
        a_1 & a_2 \\
        a_3 & a_4
    \end{bmatrix},
    \hspace{1cm}
    \vM = \begin{bmatrix}
        m_1 & 0 \\
        0 & m_4
    \end{bmatrix}
    \label{eq:deformation_matrices}
\end{equation} 
resulting in both affine and quadratic transformations. The elements of $\vA$ and $\vM$ are referred to as \textit{deformation parameters}, collected as $s = \{a_1, a_2, a_3, a_4, m_1, m_4\}$, which fully describe each geometry. We considered four types of deformations in 2D: expansion (\Gexp), shearing (\Gshear), nonlinear deformation (\Gnonlin), and rotation (\Grot), as presented in Table \ref{tab:affine_params}. Note that for the rotational family (\Grot), $\vA$ was defined as a rotation matrix parameterized by an angle $\theta$. An illustration of one representative geometry from each 2D family alongside the latent geometry is provided in Figure \ref{fig:geometries-example}.

\begin{table}
\caption{Parameter ranges for the internal ($\mathcal{G}_k$) and external ($\mathcal{G}_k^*$) families in 2D. Values were sampled uniformly from the given ranges. For \Grot{} and \GrotExt{}, $\vA$ was instead defined as a rotation matrix parameterized by an angle $\theta$.}
\label{tab:affine_params}
    \begin{center}
        \begin{small}
                \begin{tabular}{lcccc}
                \toprule
                 & $a_1, a_4$ & $a_2, a_3$ & $m_1, m_4$ & $\theta$ (degrees) \\
                \midrule
                
                \Gexp & $[1.0, 1.4]$ & $0.0$ & $0.0$ & - \\
                \GexpExt & $[1.4, 1.8]$ & $0.0$ & $0.0$ & - \\
                \midrule
                
                \Gshear & $1.0$ & $[-0.2, 0.2]$ & $0.0$ & - \\
                \GshearExt & $1.0$ & $[-0.5, -0.2] \cup [0.2, 0.5]$ & $0.0$ & - \\
                
                \midrule
                \Gnonlin & $1.0$ & $0.0$ & $[-0.015, 0.015]$ & - \\
                \GnonlinExt & $1.0$ & $0.0$ & $[-0.025, -0.015] \cup [0.015, 0.025]$ & - \\

                \midrule
                \Grot & - & - & - & $[-90, 90]$  \\
                \GrotExt & - & - & - & $\notin [-90, 90]$ \\
                
                \bottomrule
                \end{tabular}
        \end{small}
    \end{center}
  \vskip -0.1in
\end{table}

For each deformation type, we considered two separate parameter ranges. The first ``internal'' parameter range contained 50 geometries, which we used for training, validation, and within-distribution testing. We then generated 35 samples per family from the second ``external'' parameter range to test the out-of-distribution geometric generalizability of our models. In the following, we denote the \textit{external family} of geometries with a star ($^*$), as shown in Table \ref{tab:affine_params}.

In 3D, we focused exclusively on rotational deformations. Thus, the deformations were given as $\vx(s) = \vA \vX$ with $\vA \in \mathbb{R}^{3\times3}$ defined as a rotation matrix corresponding to rotations about the $x$-, $y$-, or $z$-axis. This yielded three families, each following the same internal/external split as in 2D (outlined above) with $\theta \in [-90, 90]$ for the internal families (\Hrotx, \Hroty, \Hrotz) and $\theta \notin [-90, 90]$ for the external families (\HrotxExt, \HrotyExt, \HrotzExt).

In addition to the explicit deformation parameterization given by the elements in $\vA$ and $\vM$, we also represented each geometry using a low-dimensional, data-driven descriptor obtained via \ac{PCA}. We created a separate \ac{PCA} descriptor for each internal/external pair of families (\{\Gexp, \GexpExt\}, \{\Gshear, \GshearExt\}, \{\Gnonlin, \GnonlinExt\}, \{\Grot, \GrotExt\}) in 2D and (\{\Hrotx, \HrotxExt\}, \{\Hroty,\HrotyExt\}, \{\Hrotz, \HrotzExt\}) in 3D. To derive the \ac{PCA} descriptors, we represented each geometry by a sample of 2500 uniformly distributed points, denoted by $\vx_i \in \mathbb{R}^{dN}$ across the $i$-th geometry where $d = \{2, 3\}$ is the spatial dimension and $N$ is the number of sampled points. The sampled geometries covered the full parameter ranges of the internal and external geometric families.  Next, we computed the covariance matrix defined as $\boldsymbol{C} = \sum_{i=1}^{n} (\vx_i - \Bar{\vx}) (\vx_i - \Bar{\vx})^{T} / n$ where $n = 85$ are the number of geometries used to derive a PCA model, and $\Bar{\vx}$ is the sample mean. By solving the corresponding eigenvalue problem $\boldsymbol{C}\boldsymbol{u}_k = \lambda_k \boldsymbol{u}_k$, where the eigenvectors $\boldsymbol{u}_k$ are ordered according to their corresponding eigenvalues $\lambda_1 \geq \lambda_2 \geq \cdot \cdot \cdot \geq \lambda_k$, we computed the \ac{PCA} scores as
\begin{equation}
    z_{ik} = (\vx_i - \Bar{x})^T \boldsymbol{u}_k.
\end{equation}
We normalized the scores according to $\hat{z}_{ik} = z_{ik} / \sqrt{\lambda_k}$ and used the scores associated with the first two principal components ($k = 1, 2$) as inputs to our neural networks, yielding a low-dimensional description of the geometries.

\subsection{Synthetic cardiac electrophysiology data}
\label{sec:ep_data}
We used the open source \ac{FEM} solver openCARP \citep{openCARP-paper, openCARP-sw} to create synthetic \ac{PDE} solution data for training and testing of our neural networks. To incorporate real-world conditions, these data were defined over the physical geometries $\Omega(s)$ and sparsely sampled during network training. We meshed the physical geometries using triangular elements with a maximum element size of \SI{0.01}{mm} in 2D and tetrahedral elements with a maximum edge length of 0.2 mm in 3D, and used a time step size of 0.005 ms in both 2D and 3D \cite{niederer2011verification}. The diffusion tensor $\tD$ was defined as
\begin{equation}
    \tD = \begin{bmatrix}
        \frac{\sigma_{il}\sigma_{el}}{\sigma_{il} + \sigma_{el}} & 0 & 0 \\
        0 & \frac{\sigma_{it}\sigma_{et}}{\sigma_{it} + \sigma_{et}} & 0 \\
        0 & 0 & \frac{\sigma_{in}\sigma_{en}}{\sigma_{in} + \sigma_{en}}
    \end{bmatrix}
\end{equation}
in 3D, with $\sigma_{il}, \sigma_{it}, \sigma_{in}$ providing the intracellular conductivities and $\sigma_{el}, \sigma_{et}, \sigma_{en}$ the extracellular conductivities in the longitudinal, transverse, and sheet-normal fiber directions, respectively. In 2D, a similar $\mathbb{R}^{2 \times 2}$ diffusion tensor was used with the longitudinal and transverse conductivities. In all experiments, we employed anisotropic conduction with increased conductivity along the fiber axes, mimicking healthy cardiac tissue \cite{niederer2011verification}. The fiber axes were oriented in the  $x$ direction in $\Omega_0$, and deformed according to the transformation in \eqref{eq:affine_transform} to ensure consistent dynamics in $\Omega(s)$. 

Furthermore, we initialized each \ac{PDE} solution using an external point stimulus in the center of the geometry with a radius of 0.75 mm in 2D and 1.5 mm in 3D. We ran the simulations for 620 ms and 500 ms in 2D and 3D, respectively, yielding full cycles of depolarization (activation) and repolarization. We removed data corresponding to $t < 6$ ms to exclude the applied stimulus current and to simplify physics-informed training. The exact parameter values used in the \ac{FEM} solver are listed in Table \ref{tab:opencarp_params} of Appendix \ref{appendix:synthetic_data}.

\subsection{Neural network architectures}
\label{sec:neural_networks}
\begin{figure}
    \centering
    \includegraphics[width=0.95\linewidth]{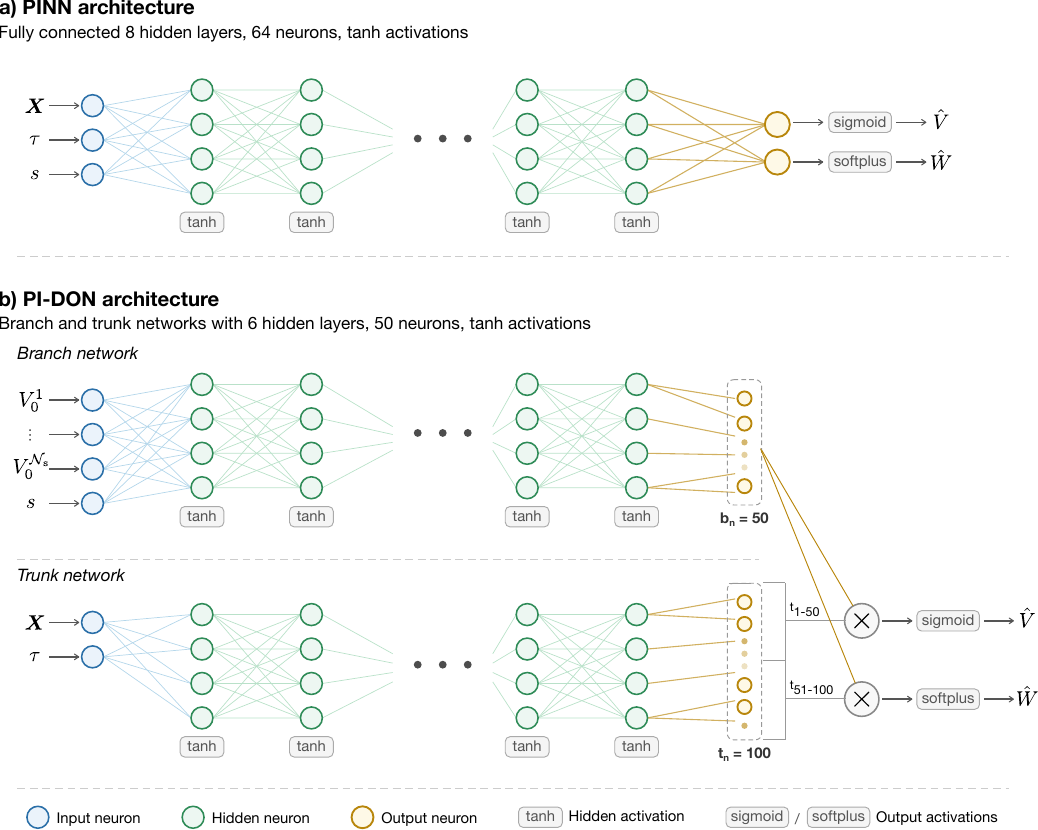}
    \caption{Schematic of the neural network architectures employed in our study. (a) The \ac{PINN} architecture consisted of a fully connected neural network containing eight hidden layers with 64 neurons in each layer. Space-time points ($\vX, \tau$) and a geometric descriptor {$s$} were given as input where applicable. (b) The \ac{PI-DON} architecture employed two sub-networks, each containing six hidden layers with 50 neurons in each layer. The branch network mapped initial condition values ($V_0$) from $\mathcal{N}_\mathrm{s}$ sensor locations for a given geometry, alongside geometric descriptors provided by $s$, to a latent representation of dimension $b_\mathrm{n} = 50$. The trunk network encoded space-time points ($\vX, \tau$) into a latent representation of dimension $t_\mathrm{n} = 100$. In both the \ac{PINN} and \ac{PI-DON} architectures, the tanh was used as activation function in the hidden layers, while the outputs were sent through the sigmoid and softplus activation functions to generate the transmembrane potential $\hat{V}$ and recovery variable $\hat{W}$, respectively.}
    \label{fig:overview-architectures}
\end{figure}

Figure \ref{fig:overview-architectures} gives an overview of the neural network architectures considered in this work, where all neural networks predicted both $\hat{V}$ and $\hat{W}$ as outputs. Additionally, since $V\in[0,1]$ and $W\in[0,\infty]$ in \eqref{eq:mapped_ap}, the outputs of the final hidden network layer were summed up and sent through the sigmoid and softplus activation functions, respectively, in both architectures. In this way, we ensured that the predicted outputs were in the correct range, thereby avoiding large gradients during training initialization. In the remaining parts of the neural networks, we employed the tanh activation function \cite{wang2023expert} to handle the second-order derivatives in \eqref{eq:mapped_ap} needed to calculate the physics loss in \eqref{eq:mse_R}. 

Furthermore, each \ac{PINN} \cite{raissi2019physics, wang2023expert} comprised a fully connected network with eight hidden layers, with 64 neurons in each layer (Figure \ref{fig:overview-architectures}a), resulting in 33,280 trainable parameters in all hidden layers. The inputs comprised space-time points, as well as geometric descriptors for the given geometry, where applicable. An overview of the hyperparameters used in the \acp{PINN} is provided in Table \ref{tab:models-config-pinns} of Appendix \ref{appendix:model_configs}.

The \ac{PI-DON}s \cite{wang2021learning} employed separate branch and trunk networks (Figure \ref{fig:overview-architectures}b), each with six hidden layers of 50 neurons per layer resulting in 30,601 trainable parameters in the hidden layers. Additionally, the trunk network had an output layer of 100 neurons, while the branch network had an output layer of 50 neurons, inspired by the architecture used to solve the reaction-diffusion equation, see previous work by \citet{wang2021learning}. Thus, the branch outputs were dotted twice with 50 components of the trunk outputs, yielding two scalar outputs corresponding to $\hat{V}$ and $\hat{W}$. Only the trunk network was endowed with additional neurons in its last output layer to support the representation of two outputs, since $V$ and $W$ in \eqref{eq:mapped_ap} are closely related. Moreover, the trunk network received space-time points sampled across the input geometry, whereas the branch network received the initial condition ($V_0$) as inputs at fixed sensor locations together with the geometric descriptors describing the geometry. In 2D, the number of sensor locations were equal to the number of observed data measurements for the \ac{PINN} ($\mathcal{N}_{\mathrm{s}} = \mathcal{N}_{data} = 14$), while in 3D, the number of sensor locations and observed data measurements were increased to $\mathcal{N}_{\mathrm{s}} = \mathcal{N}_{data} = 42$ for the \ac{PI-DON}. Furthermore, we used an interpolation scheme to compute the transmembrane potential $V$ at the fixed sensor locations. An overview of the hyperparameter settings for the \acp{PI-DON} is provided in Table \ref{tab:models-config-deeponet} of Appendix~\ref{appendix:model_configs}.

\subsection{Ablation models, training procedures and evaluation methods}
\label{sec:training_procedures}
We developed a collection of \acp{PINN} \cite{raissi2019physics} and \acp{PI-DON} \cite{wang2021learning} to assess and compare the latent \ac{PDE} mapping technique (Table~\ref{tab:pinns-overview}). Models denoted by LPM employed our proposed latent PDE mapping framework, whereas ablated models systematically removed key components. In particular, models marked with LG incorporated a latent geometry for the input data but lacked the pull-back~\eqref{eq:generic-pde-mapped} of the \ac{PDE} and \ac{BC} residuals in the physics loss. The \ac{PA-PINN} defined its variables with respect to the physical geometries, without access to the latent geometry $\Omega_0$. The \ac{PA-PINN} nevertheless received shape parameters as input, either as deformation parameters or PCA components. Finally, \PINN{} lacked access to both the latent geometry and the shape parameters. The \acp{PI-DON} (\LPMDeepO, \LGDeepO) were restricted to inputs from $\Omega_0$, to ensure fixed sensor locations as proposed by \citet{lu2019deeponet}.

\begin{table}[tb]
  \caption{Overview of physics-informed models used in our experiments. Models marked with LPM apply our proposed framework, whereas models denoted by LG take data from the latent geometry $\Omega_0$ and shape parameters as inputs. The \ac{PA-PINN} and Basic-PINN trained directly on the physical geometries, while the \ac{PA-PINN} additionally incorporates shape parameters. The \acp{PI-DON} (\LPMDeepO, \LGDeepO) were restricted to inputs from $\Omega_0$, to ensure fixed sensor locations \cite{lu2019deeponet}.}
  \label{tab:pinns-overview}
  \begin{center}
    \begin{small}
      \begin{sc}
        \begin{tabular}{lcccc}
          \toprule
            & Input augmented with shape parameters  & Input data from $\Omega_0$ & Latent PDE mapping  \\
          \midrule
          \LPM    & $\surd$ & $\surd$ & $\surd$  \\
          \LPMDeepO    & $\surd$ & $\surd$ & $\surd$  \\
          \LG & $\surd$ & $\surd$ & \\
          \LGDeepO & $\surd$ & $\surd$ & \\
          \Affine    & $\surd$ &  &  \\
          \PINN    &  & &  \\
          \bottomrule
        \end{tabular}
      \end{sc}
    \end{small}
  \end{center}
  \vskip -0.1in
\end{table}

All neural networks were trained by minimizing a hybrid loss function defined as \begin{equation}
    \mathcal{L}(\theta) = \mathcal{L}_{data}(\theta) + \mathcal{L}_{phys}(\theta) + \mathcal{L}_{bc}(\theta) + \mathcal{L}_{ic}(\theta)
\end{equation}
where $\mathcal{L}_{data}(\theta)$ is the discrepancy to the \ac{FEM} data, $\mathcal{L}_{phys}(\theta)$ is the loss described by the governing \ac{PDE} residual, $\mathcal{L}_{bc}(\theta)$ is the loss associated with the boundary condition, and $\mathcal{L}_{ic}(\theta)$ is the loss associated with the initial condition. The loss terms were equally weighted, and each term was evaluated using the \ac{MSE} sampled from the \ac{FEM} space-time discretization points $\mathcal{N}_{data}$, $\mathcal{N}_{phys}$, $\mathcal{N}_{bc}$, $\mathcal{N}_{ic}$ (Appendix Table \ref{tab:models-config-colloc} ). Furthermore, we defined the physics and boundary losses in the LPM models as
\begin{equation}
    \mathcal{L}_{phys}^{LPM} = \frac{1}{\mathcal{N}_{phys}} \sum_{i}^{\mathcal{N}_{phys}} \left[ \mathcal{R}^{LPM}(\vX, \tau, \hat{V}, \hat{W}, \tF, J; s) \right]^2, \hspace{1cm}
    \mathcal{L}_{bc}^{LPM} = \frac{1}{\mathcal{N}_{bc}} \sum^{\mathcal{N}_{bc}} \left[ \mathcal{R}^{LPM}(\vX, \tau, \hat{V}, \tF, J; s) \right]^2
    \label{eq:L-phys-LPM}
    \end{equation}
and the conventional losses in the ablated models as
\begin{equation}
    \mathcal{L}_{phys}^{conv} = \frac{1}{\mathcal{N}_{phys}} \sum^{\mathcal{N}_{phys}} \left[ \mathcal{R}(\vx, \tau, \hat{V}, \hat{W}) \right]^2, \hspace{1cm}
    \mathcal{L}_{bc}^{conv} = \frac{1}{\mathcal{N}_{bc}} \sum^{\mathcal{N}_{bc}} \left[ \mathcal{R}(\vx, \tau, \hat{V}) \right]^2,
    \end{equation}
where $\vX \in \Omega_0$ and $\vx \in \Omega(s)$. The deformation gradient $\tF$ and Jacobian $J$ in \eqref{eq:L-phys-LPM} were precomputed for each geometry and treated as constant coefficient fields during training. Thus, the shape gradient ${\partial \mathcal{L}_{phys}}/{\partial s}$ focused on how the solution changes with the geometry rather than how the deformation map changes with the geometry, as highlighted in Section \ref{app:computational-graph} of the Appendix.

Furthermore, we split each internal family in $\mathcal{G}_k$ and $\mathcal{H}_{rot}^k$ into a training set, a validation set, and a test set. Unlike the conventional split used in machine learning, we adopted an inverted allocation strategy with $20\%$ train data, $10\%$ validation data, and $70\%$ test data to restrict the available training data. Thus, each family ($\mathcal{G}_k$, $\mathcal{H}_{rot}^k$) had 10 train geometries, 5 validation geometries, and 35 test geometries.

During training, we evaluated $\mathcal{L}(\theta)$ for each geometry in the validation set. Since our validation set spanned multiple distinct geometries, we selected the best model state as that which gave the lowest maximum $\mathcal{L}(\theta)$ across the validation geometries, rather than the lowest average $\mathcal{L}(\theta)$. This criterion encouraged generalization to geometries that differ substantially from those seen during training. To reduce computational cost, we computed the validation loss every 10 epochs using a factor 100 downsampling of the \ac{FEM} space-time discretization points.

To evaluate PDE solution accuracy, we used the relative $L_2$ error, given as
\begin{equation}
    e_{L^2} = \frac{ \sqrt{ \sum_{i}^{\mathcal{N}_{test}} \left( \hat{V}_i - V_i \right)^{2}}} {\sqrt{ \sum_i^{\mathcal{N}_{test}}  V_i ^{2}}},
\label{eq:rmse}
\end{equation}
where $\hat{V}$ is the predicted transmembrane potential and $V$ originates from the \ac{FEM} solution. $\mathcal{N}_{test}$ includes all space-time points from a given geometry. In our results, we present the mean relative $L_2$ error across test geometries $\overline{e}_{L^2}$, 
along with the corresponding standard deviation. 

Additionally, we assessed statistical significance between the best and second-best performing models using a paired Wilcoxon signed-rank test \cite{wilcoxon1945individual} at a 5\% significance level. We chose this test because it is designed for paired observations, allowing a direct comparison of model performance on the same test geometries, while not requiring the paired differences to follow a normal distribution. The test was performed on per-geometry $e_{L^2}$, treating each geometry as a paired observation. Results marked with * indicate a statistically significant difference.

\section{Results}
\label{sec:experiments}
In the following sections, we present results from a series of experiments used to assess the latent \ac{PDE} mapping technique and its applicability.

\subsection{Latent PDE mapping improves geometric generalizability with limited training data}
\label{sec:2d-results}
Table \ref{tab:2d-single-deformations} reports the mean relative $L_2$ error ($\Bar{e}_{L^2}$) for 16 sets of experiments in 2D, wherein each experiment consisted of a geometric family and a geometry descriptor. The results showed that the \LPMDeepO{} outperformed the \LGDeepO{} in 14 out 16 experiments, while the \LPM{} outperformed the ablation \ac{PINN} models in 11 out of 16 experiments. Out of these results, 13 and 9 experiments yielded significantly lower errors with the \LPMDeepO{} and \LPM, respectively, when comparing the results with the second-best performing models.

\begin{figure}
  \begin{center}
    \centerline{\includegraphics[width=\textwidth]{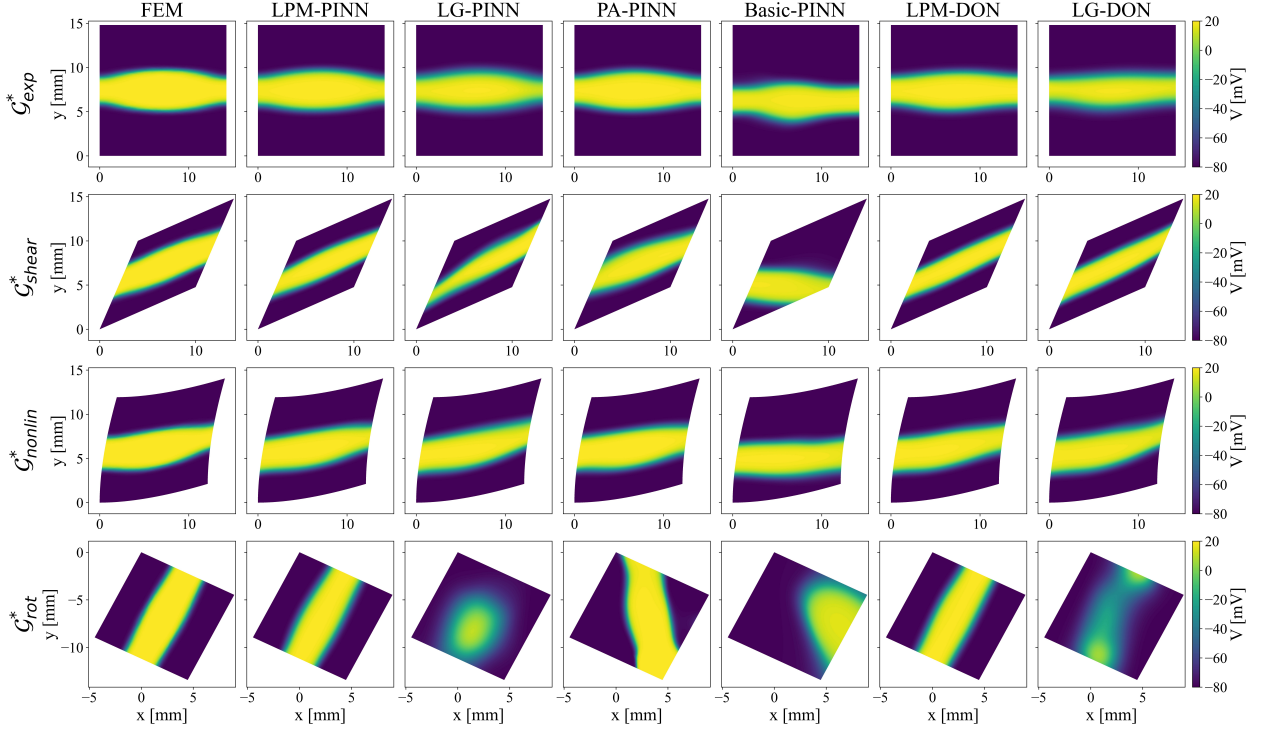}}
    \caption{
      Snapshots of predicted transmembrane voltages ($V$) at $t = 50$ ms when using deformation parameters as geometric descriptors. Each row corresponds to a geometry taken from the presented external family (\GexpExt, \GshearExt, \GnonlinExt, \GrotExt). The left column shows the FEM ground truth approximation. An animated version of the figure is available in the Supplementary material.
    }
    \label{fig:2d-single-deformation-results}
  \end{center}
\end{figure}

To help contextualize our results, we visualized solutions snapshots at $t = 50$ ms for the experiments involving the external geometries (Figure~\ref{fig:2d-single-deformation-results}). The visualizations show that the LPM models yielded plausible solutions that visibly matched the \ac{FEM} data, whereas the ablation models contained noticeable inconsistencies and errors, especially for the rotation families, \Grot{} and \GrotExt. Indeed, out of the four types of geometry families (expansion, shear, nonlinear, rotation), the advantage of the LPM technique was most apparent for the rotation families \Grot{} and \GrotExt. On \GrotExt, the second-best performing model \LG{} achieved $\Bar{e}_{L^2} = 0.289$, whereas \LPM{} achieved $\Bar{e}_{L^2} = 0.068$, corresponding to a $4.3\times$ improvement in error. Additionally, Figure \ref{fig:2d-single-deformation-results} shows that the higher error for \LG{} on the rotation family \GrotExt{} was associated with a visibly incorrect solution, in which the prediction resembled a localized point-like response, while \LPM{} preserved the correct wave propagation pattern as seen in the \ac{FEM} solution. This is further observable in the animated version of Figure \ref{fig:2d-single-deformation-results} in the Supplementary material, where \LG{} produced a notably incorrect solution across the entire time trajectory, while \LPM{} accurately captured both the depolarization and repolarization dynamics across the geometry.

\begin{table}
    \caption{Mean relative $L_2$ prediction-FEM discrepancy ($\overline{e}_{L^2}$) $\pm$ std evaluated over the internal ($\mathcal{G}_k$) and external ($\mathcal{G}_k^*$) test geometries in 2D. Best performing models for each neural network architecture are given in \textbf{bold}, while values marked with * indicate statistically significant differences (at the 5\% level) according to a paired Wilcoxon signed-rank test \cite{wilcoxon1945individual} between the best and second-best performing models. Models marked with LPM utilize the latent \ac{PDE} mapping framework. The geometrical descriptor of each experiment is written in vertical text in the leftmost column.}
    \label{tab:2d-single-deformations}
    \begin{center}
        \begin{small}
            \begin{sc}
            \hspace*{-0.8cm}
            \begin{tabular}{llcccccc}

             & \multicolumn{5}{c}{Fully Connected Networks} & \multicolumn{2}{c}{DeepOnets} \\
            \cmidrule[\heavyrulewidth](lr){2-6}
            \cmidrule[\heavyrulewidth](lr){7-8}
            &\multicolumn{1}{c}{}  &\multicolumn{1}{c}{\LPM} &\multicolumn{1}{c}{\LG} &\multicolumn{1}{c}{\Affine} &\multicolumn{1}{c}{\PINN} &\multicolumn{1}{c}{\LPMDeepO} &\multicolumn{1}{c}{\LGDeepO} \\
            
            \cmidrule(lr){2-6}
            \cmidrule(lr){7-8}
            
            \multirow{8}{*}{\rotatebox{90}{Deformation}} & \Gexp & \hspace{0.16cm}\textbf{0.050 $\pm$ 0.005} & 0.066 $\pm$ 0.008 & 0.050 $\pm$ 0.005 & 0.201 $\pm$ 0.078 & *\textbf{0.093 $\pm$ 0.012} & \hspace{0.16cm}0.113 $\pm$ 0.018 \\
            & \GexpExt & \hspace{0.16cm}0.082 $\pm$ 0.019 & 0.115 $\pm$ 0.018 & \textbf{0.080 $\pm$ 0.015} & 0.491 $\pm$ 0.102 & \hspace{0.16cm}0.249 $\pm$ 0.067 & *\textbf{0.241 $\pm$ 0.049} \\
            & \Gshear & *\textbf{0.049 $\pm$ 0.007} & 0.059 $\pm$ 0.012 & 0.051 $\pm$ 0.009 & 0.228 $\pm$ 0.078 & *\textbf{0.082 $\pm$ 0.006} & \hspace{0.16cm}0.090 $\pm$ 0.007 \\
            & \GshearExt & \hspace{0.16cm}\textbf{0.130 $\pm$ 0.062} & 0.153 $\pm$ 0.055 & 0.137 $\pm$ 0.037 & 0.481 $\pm$ 0.076 & *\textbf{0.138 $\pm$ 0.041} & \hspace{0.16cm}0.149 $\pm$ 0.040 \\
            & \Gnonlin & *\textbf{0.052 $\pm$ 0.006} & 0.057 $\pm$ 0.013 & 0.063 $\pm$ 0.007 & 0.154 $\pm$ 0.040 & \hspace{0.16cm}0.086 $\pm$ 0.007 & *\textbf{0.082 $\pm$ 0.014} \\
            & \GnonlinExt & *\textbf{0.086 $\pm$ 0.022} & 0.100 $\pm$ 0.028 & 0.096 $\pm$ 0.027 & 0.290 $\pm$ 0.086 & \hspace{0.16cm}\textbf{0.117 $\pm$ 0.031} & \hspace{0.16cm}0.120 $\pm$ 0.034 \\
            & \Grot & *\textbf{0.043 $\pm$ 0.002} & 0.145 $\pm$ 0.042 & 0.083 $\pm$ 0.031 & 0.413 $\pm$ 0.050 & *\textbf{0.078 $\pm$ 0.002} & \hspace{0.16cm}0.199 $\pm$ 0.029 \\
            & \GrotExt & *\textbf{0.068 $\pm$ 0.018} & 0.289 $\pm$ 0.027 & 0.578 $\pm$ 0.158 & 0.686 $\pm$ 0.124  & *\textbf{0.077 $\pm$ 0.002} & \hspace{0.16cm}0.235 $\pm$ 0.002 \\
            
            \cmidrule(lr){2-6}
            \cmidrule(lr){7-8}
            
            \multirow{8}{*}{\rotatebox{90}{PCA}} & \Gexp & \hspace{0.16cm}0.056 $\pm$ 0.006 & 0.070 $\pm$ 0.013 & *\textbf{0.054 $\pm$ 0.007} & 0.201 $\pm$ 0.078 & *\textbf{0.091 $\pm$ 0.012} & \hspace{0.16cm}0.110 $\pm$ 0.018 \\
            & \GexpExt &  \hspace{0.16cm}0.125 $\pm$ 0.040 & 0.146 $\pm$ 0.030 & *\textbf{0.113 $\pm$ 0.031} & 0.491 $\pm$ 0.102 & *\textbf{0.252 $\pm$ 0.072} & \hspace{0.16cm}0.263 $\pm$ 0.064 \\
            & \Gshear & *\textbf{0.049 $\pm$ 0.008} & 0.067 $\pm$ 0.013 & \hspace{0.16cm}0.052 $\pm$ 0.009 & 0.228 $\pm$ 0.078 & *\textbf{0.082 $\pm$ 0.006} & \hspace{0.16cm}0.092 $\pm$ 0.006 \\
            & \GshearExt & \hspace{0.16cm}0.142 $\pm$ 0.087 & \hspace{0.16cm}0.172 $\pm$ 0.082 & \hspace{0.16cm}\textbf{0.139 $\pm$ 0.051} & 0.481 $\pm$ 0.076 & *\textbf{0.137 $\pm$ 0.042} & \hspace{0.16cm}0.148 $\pm$ 0.038 \\
            & \Gnonlin & *\textbf{0.058 $\pm$ 0.012} & 0.064 $\pm$ 0.018 & \hspace{0.16cm}0.061 $\pm$ 0.009 & 0.154 $\pm$ 0.040 & *\textbf{0.081 $\pm$ 0.008} & \hspace{0.16cm}0.090 $\pm$ 0.013 \\
            & \GnonlinExt & \hspace{0.16cm}0.114 $\pm$ 0.035 & 0.121 $\pm$ 0.055 & *\textbf{0.105 $\pm$ 0.030} & 0.290 $\pm$ 0.086 & *\textbf{0.118 $\pm$ 0.035} & \hspace{0.16cm}0.137 $\pm$ 0.042 \\
            & \Grot & *\textbf{0.044 $\pm$ 0.003} & 0.152 $\pm$ 0.038 & \hspace{0.16cm}0.079 $\pm$ 0.025 & 0.413 $\pm$ 0.050 & *\textbf{0.076 $\pm$ 0.002} & \hspace{0.16cm}0.193 $\pm$ 0.030 \\
            & \GrotExt & *\textbf{0.060 $\pm$ 0.010} & 0.307 $\pm$ 0.051 & \hspace{0.16cm}0.586 $\pm$ 0.188 & 0.686 $\pm$ 0.124  & *\textbf{0.076 $\pm$ 0.002} & \hspace{0.16cm}0.237 $\pm$ 0.003 \\

            \cmidrule[\heavyrulewidth](lr){2-6}
            \cmidrule[\heavyrulewidth](lr){7-8}
            
            \end{tabular}
            \end{sc}
        \end{small}
    \end{center}
\end{table}

The bottom panel in Table \ref{tab:2d-single-deformations} presents the results when using \ac{PCA} scores instead of deformation parameters as geometric descriptors. Across all models, the largest performance differences remained on the rotational families \Grot{} and \GrotExt{}. For the \acp{PI-DON}, \LPMDeepO{} yielded significantly lower prediction errors than \LGDeepO{} across all deformation families. Likewise, \LPM{} achieved significantly lower prediction errors ($\Bar{e}_{L^2} = 0.044$ and $0.060$) than the second-best performing model \LG \ ($\Bar{e}_{L^2} = 0.152$ and $0.307$) on the rotational families \Grot{} and \GrotExt{}, respectively.

\subsection{Quantification of missing boundary shape gradients without latent \ac{PDE} mapping}
\begin{figure}
  \begin{center}
    \centerline{\includegraphics[width=\textwidth]{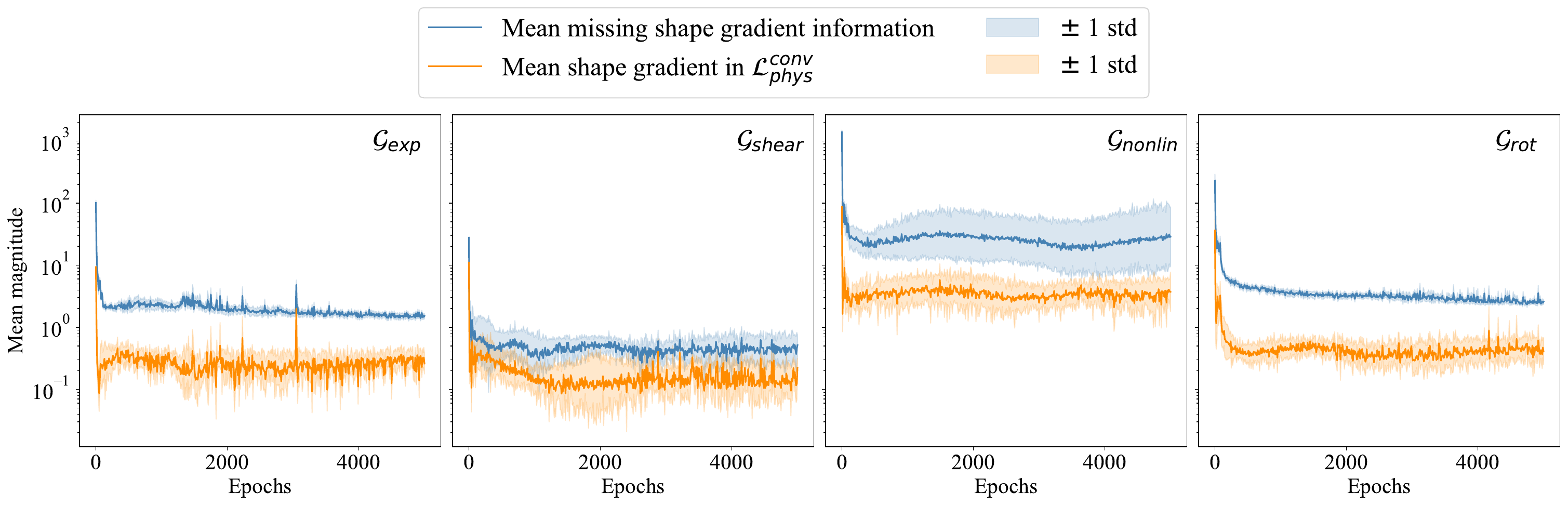}}
    \caption{
      Quantification of the mean shape gradient and mean missing boundary shape gradient, together with their corresponding standard deviations, computed across training geometries in the given two-dimensional families when training the \Affine{} with deformation parameters as geometric descriptor. 
    }
    \label{fig:2d-missing-shape-gradients}
  \end{center}
\end{figure}

We performed a numerical approximation of \eqref{eq:reynold}, as described in Appendix \ref{appendix:numerical_analysis}, to inspect the magnitude of the shape gradient $\partial \mathcal{L}_{\text{phys}}^{\text{conv}} / \partial s$ encountered during training of \Affine{}, as well as the magnitude of the missing shape gradient induced by boundary motion, corresponding to the second term in \eqref{eq:reynold}. Figure \ref{fig:2d-missing-shape-gradients} compares the magnitudes of these two terms for each deformation family. Across all families, the boundary-motion contribution (blue) was larger than the shape gradient used in \Affine{} (orange). The largest gradient magnitudes were observed for \Gnonlin{} and \Grot{}, whereas smaller magnitudes were obtained for \Gexp{} and \Gshear{}.

These observations coincided with the performance reported in Table \ref{tab:2d-single-deformations}. The largest differences between the latent \ac{PDE} mapping models and their corresponding ablation models were observed for \Gnonlin{} and \Grot{}, which were also the families exhibiting the largest boundary-motion contributions in Figure \ref{fig:2d-missing-shape-gradients}. Nevertheless, it should be emphasized that the gradient size shown in Figure \ref{fig:2d-missing-shape-gradients} did not necessarily correspond to the scale of performance improvements, as neural network training is a complex process influenced by multiple, interacting factors.

\subsection{Latent PDE mapping extends to three-dimensional geometries}
Table \ref{tab:3d-single-deformations} reports the mean relative $L_2$ error ($\Bar{e}_{L^2}$) for 12 experiments in 3D, wherein each experiment consisted of a rotational family and a geometry descriptor. The results showed that the \LPMDeepO{} outperformed the \LGDeepO{} in 8 out 12 experiments, while the \LPM{} outperformed the ablation \ac{PINN} models in 12 out of 12 experiments. Out of these results, 8 and 11 experiments yielded significantly lower errors with the \LPMDeepO{} and \LPM, respectively, when comparing the results with the second-best performing models.

\begin{figure}
  \begin{center}
    \centerline{\includegraphics[width=\textwidth]{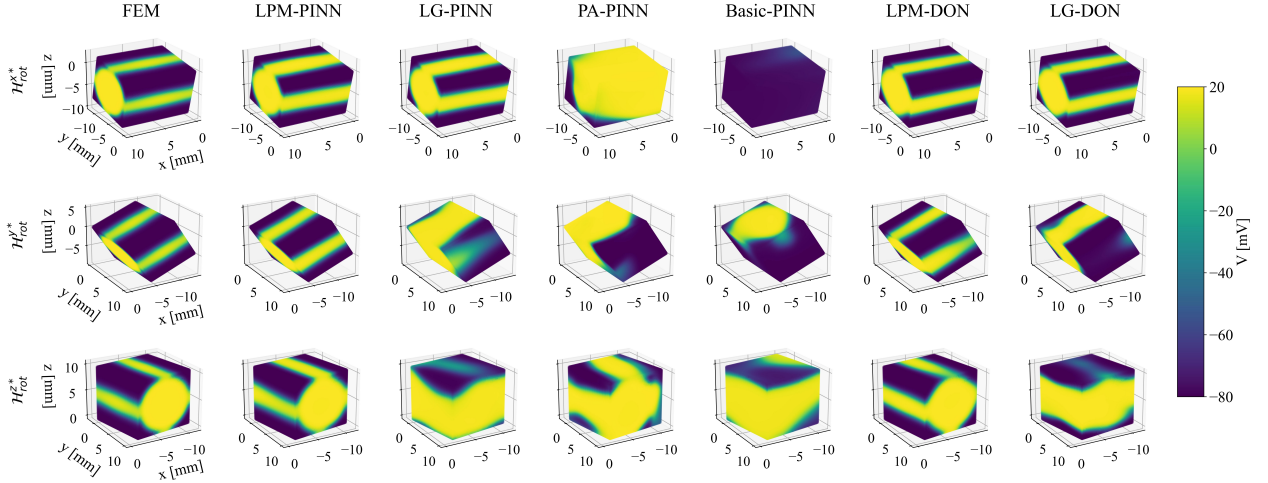}}
    \caption{
      Snapshots of predicted transmembrane voltages ($V$) at $t = 100$ ms when using deformation parameters as geometric descriptors. Each row corresponds to a geometry taken from the presented external family (\HrotxExt, \HrotyExt, \HrotzExt). The left column shows the FEM ground truth approximation. An animated version of the figure is available in the Supplementary material.
    }
    \label{fig:3d-single-deformation-results}
  \end{center}
\end{figure}

Our results were further contextualized by Figure \ref{fig:3d-single-deformation-results}, which visualizes solutions snapshots at $t = 100$ ms for the external geometries. The visualization shows that the LPM models accurately captured wave dynamics, closely matching the \ac{FEM} data across all experiments. The improvements were most notable for \HrotyExt{} and \HrotzExt{}, where the LPM models appeared to yield accurate solutions, whereas the ablation models made visible errors and included solution inconsistencies. This trend aligned with the degree of geometric distortion in the \ac{PDE} solutions; the geometric deformations of \HrotyExt{} and \HrotzExt{} rotated the axis of anisotropy (i.e. fiber direction), whereas the deformations in \HrotxExt{} did not. For \HrotyExt{} and \HrotzExt{}, \LPM{} attained $\bar{e}_{L^2} =$ 0.054 and 0.093, compared with $\bar{e}_{L^2} =$ 0.273 and 0.363 for the second-best performing model \LG{}, corresponding to $5.1\times$ and $3.9\times$ error reductions, respectively. Additionally, \LPMDeepO{} achieved $\bar{e}_{L^2}=$ 0.086 and 0.079 on \HrotyExt{} and \HrotzExt{}, compared to $\bar{e}_{L^2} =$ 0.222 and 0.258 for the second-best performing model \LGDeepO{}, yielding $2.6\times$ and $3.3\times$ error improvements. These error improvements were apparent across the entire time trajectory in the animated version of Figure \ref{fig:3d-single-deformation-results} (Supplementary material). Here, the LPM models were able to accurately capture wave dynamics closely resembling the \ac{FEM} data across all time points, whereas the ablation models showed visible errors and inconsistencies in predictions.

The bottom panel of Table \ref{tab:3d-single-deformations} shows that replacing deformation parameters with \ac{PCA} scores yielded similar behavior across all models. In particular, \Hroty, \HrotyExt, \Hrotz{} and \HrotzExt{} remained the most challenging cases, where \LPM{} and \LPMDeepO{} consistently achieved significantly lower prediction errors than their corresponding ablation models.

\begin{table}
    \caption{Mean relative $L_2$ prediction-FEM discrepancy ($\overline{e}_{L^2}$) $\pm$ std evaluated over the internal ($\mathcal{H}_{rot}^k$) and external ($\mathcal{H}_{rot}^{k*}$) test geometries in 3D. Best performing models for each neural network architecture are given in \textbf{bold}, while values marked with * indicate statistically significant differences (at the 5\% level) according to a paired Wilcoxon signed-rank test \cite{wilcoxon1945individual} between the best and second-best performing models. Models marked with LPM utilize the latent \ac{PDE} mapping framework. The geometrical descriptor of each experiment is written in vertical text in the leftmost column.}
    \label{tab:3d-single-deformations}
    \begin{center}
        \begin{small}
            \begin{sc}
            \begin{tabular}{llcccccc}

             & \multicolumn{5}{c}{Fully Connected Networks} & \multicolumn{2}{c}{DeepOnets} \\
            \cmidrule[\heavyrulewidth](lr){2-6}
            \cmidrule[\heavyrulewidth](lr){7-8}
            &\multicolumn{1}{c}{}  &\multicolumn{1}{c}{\LPM} &\multicolumn{1}{c}{\LG} &\multicolumn{1}{c}{\Affine} &\multicolumn{1}{c}{\PINN} &\multicolumn{1}{c}{\LPMDeepO} &\multicolumn{1}{c}{\LGDeepO} \\
            
            \cmidrule(lr){2-6}
            \cmidrule(lr){7-8}
            
            \multirow{6}{*}{\rotatebox{90}{Deformation}} & \Hrotx & \hspace{0.16cm}\textbf{0.052 $\pm$ 0.003} & 0.052 $\pm$ 0.003 & 0.110 $\pm$ 0.023 & 0.430 $\pm$ 0.043 & \hspace{0.16cm}0.104 $\pm$ 0.001 & *\textbf{0.094 $\pm$ 0.001} \\
            & \HrotxExt & *\textbf{0.072 $\pm$ 0.008} & 0.073 $\pm$ 0.009 & 0.356 $\pm$ 0.116 & 1.027 $\pm$ 0.294 & \hspace{0.16cm}0.104 $\pm$ 0.001 & *\textbf{0.094 $\pm$ 0.001} \\
            & \Hroty & *\textbf{0.040 $\pm$ 0.001} & 0.151 $\pm$ 0.028 & 0.120 $\pm$ 0.025 & 0.365 $\pm$ 0.041 & *\textbf{0.086 $\pm$ 0.001} & \hspace{0.16cm}0.197 $\pm$ 0.019 \\
            & \HrotyExt & *\textbf{0.054 $\pm$ 0.007} & 0.273 $\pm$ 0.032 & 0.593 $\pm$ 0.209 & 0.638 $\pm$ 0.140 & *\textbf{0.086 $\pm$ 0.001} & \hspace{0.16cm}0.222 $\pm$ 0.001 \\
            & \Hrotz & *\textbf{0.047 $\pm$ 0.004} & 0.164 $\pm$ 0.037 & 0.112 $\pm$ 0.030 & 0.366 $\pm$ 0.025 & *\textbf{0.078 $\pm$ 0.001} & \hspace{0.16cm}0.240 $\pm$ 0.013 \\
            & \HrotzExt & *\textbf{0.093 $\pm$ 0.023} & 0.363 $\pm$ 0.076 & 0.557 $\pm$ 0.205 & 0.456 $\pm$ 0.040 & *\textbf{0.079 $\pm$ 0.001} & \hspace{0.16cm}0.258 $\pm$ 0.001 \\
            
            \cmidrule(lr){2-6}
            \cmidrule(lr){7-8}
            
            \multirow{6}{*}{\rotatebox{90}{PCA}} & \Hrotx & *\textbf{0.053 $\pm$ 0.004} & 0.056 $\pm$ 0.003 & 0.102 $\pm$ 0.018 & 0.430 $\pm$ 0.043 & \hspace{0.16cm}0.079 $\pm$ 0.000 & *\textbf{0.070 $\pm$ 0.001} \\
            & \HrotxExt &  *\textbf{0.076 $\pm$ 0.008} & 0.079 $\pm$ 0.009 & 0.335 $\pm$ 0.099 & 1.027 $\pm$ 0.294 & \hspace{0.16cm}0.078 $\pm$ 0.001 & *\textbf{0.070 $\pm$ 0.001} \\
            & \Hroty & *\textbf{0.040 $\pm$ 0.001} & 0.158 $\pm$ 0.031 & 0.109 $\pm$ 0.021 & 0.365 $\pm$ 0.041 & *\textbf{0.073 $\pm$ 0.001} & \hspace{0.16cm}0.198 $\pm$ 0.019 \\
            & \HrotyExt & *\textbf{0.080 $\pm$ 0.023} & 0.291 $\pm$ 0.039 & 0.499 $\pm$ 0.144 & 0.638 $\pm$ 0.140 & *\textbf{0.073 $\pm$ 0.001} & \hspace{0.16cm}0.223 $\pm$ 0.001 \\
            & \Hrotz & *\textbf{0.049 $\pm$ 0.007} & 0.156 $\pm$ 0.043 & 0.128 $\pm$ 0.033 & 0.366 $\pm$ 0.025 & *\textbf{0.082 $\pm$ 0.001} & \hspace{0.16cm}0.237 $\pm$ 0.016 \\
            & \HrotzExt & *\textbf{0.112 $\pm$ 0.033} & 0.303 $\pm$ 0.037 & 0.576 $\pm$ 0.187 & 0.456 $\pm$ 0.040 & *\textbf{0.082 $\pm$ 0.001} & \hspace{0.16cm}0.262 $\pm$ 0.002 \\

            \cmidrule[\heavyrulewidth](lr){2-6}
            \cmidrule[\heavyrulewidth](lr){7-8}
            
            \end{tabular}
            \end{sc}
        \end{small}
    \end{center}
\end{table}

\subsection{Effect of latent PDE mapping on computational cost}
\begin{figure}
  \begin{center}
    \centerline{\includegraphics[width=\textwidth]{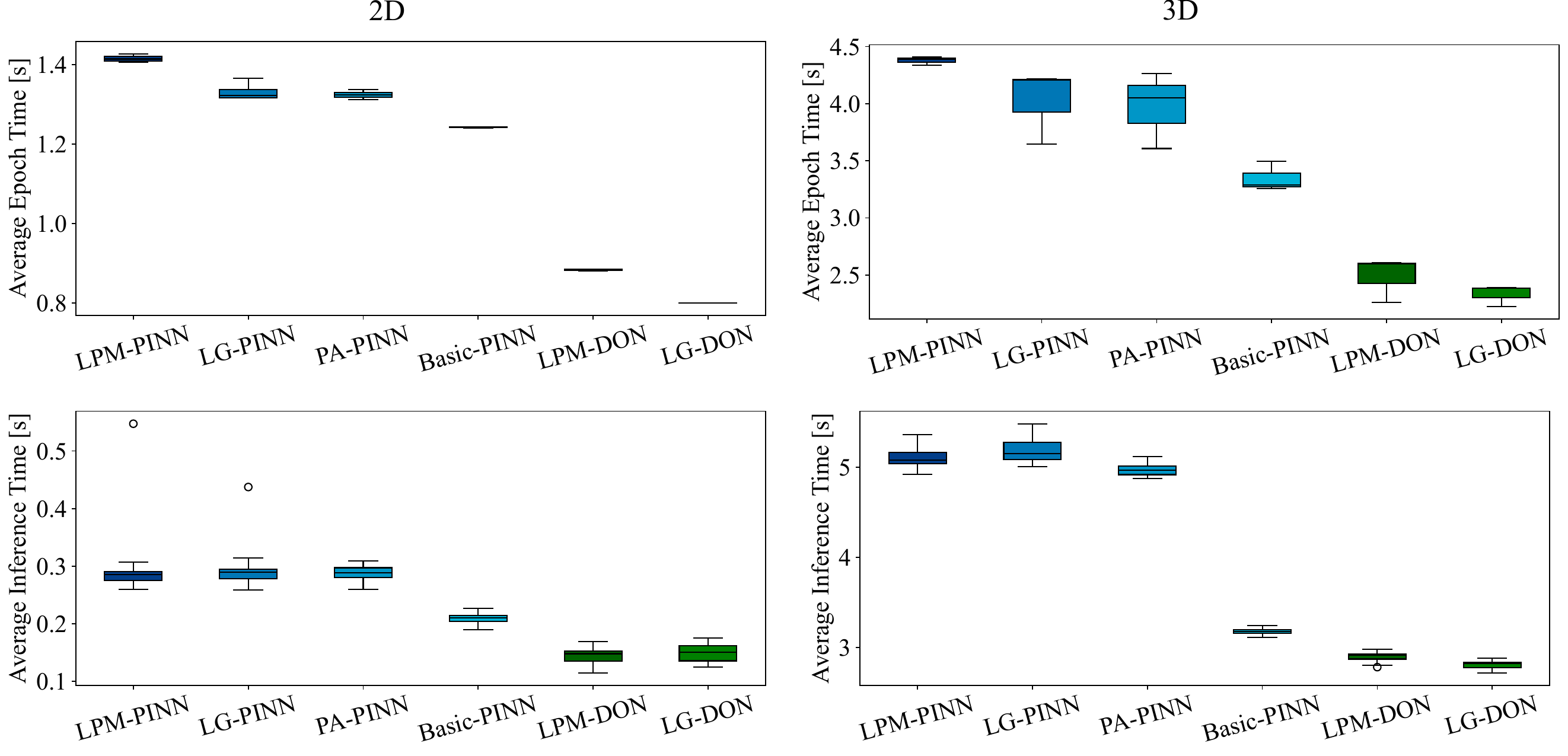}}
    \caption{
      Boxplots of the average training time per epoch in 2D and 3D (top row) and the average inference time per geometry in 2D and 3D (bottom row). Due to variability in system load of the shared computing environment, training times were averaged across all epochs for each geometry family. Additionally, due to differences in space-time points used for evaluation in each geometry family, inference times were averaged across corresponding geometry indices within each family (e.g., geometry 1 across \Gexp, \GexpExt, \Gshear, \GshearExt, \Gnonlin, \GnonlinExt, \Grot, and \GrotExt{} in the 2D experiments), to reduce variability between geometric families.  All reported experiments used deformation parameters as geometric descriptors and were run on an NVIDIA HGX H200 GPU in a shared computing environment.
    }
    \label{fig:computational-times}
  \end{center}
\end{figure}

Figure \ref{fig:computational-times} presents the training and inference times for the \acp{PINN} and \acp{PI-DON} in both 2D and 3D. Across all models, the introduction of latent \ac{PDE} mapping resulted in a small increase in the computational cost per training epoch. This increase coincided with the additional evaluation of the deformation gradient $\tF$ within the physics loss. The average time required to compute $\tF$ during mesh generation was $22.5 \pm 10.6$ s and $318.4 \pm 36.3$ s per geometry in 2D and 3D, respectively. In the present implementation, $\tF$ was precomputed during mesh generation and reused during training, avoiding repeated evaluation.

Furthermore, Figure \ref{fig:computational-times} shows comparable inference times across \LPM, \LG, and \Affine, as well as for \LPMDeepO{} and \LGDeepO{}. A somewhat lower inference time was observed for the \PINN, most likely due to the reduced number of input parameters for this specific model. Overall, the results demonstrated that the additional computational cost introduced by latent \ac{PDE} mapping was modest and primarily confined to the training phase.

\section{Discussion}
\label{sec:discussion}
In this study, we introduced latent \ac{PDE} mapping (LPM), a physics-informed learning framework designed to enable consistent training across diverse geometries. Our results in Tables \ref{tab:2d-single-deformations} and \ref{tab:3d-single-deformations} demonstrated that LPM improved solution accuracy in the majority of the experimental scenarios. These trends were further confirmed by visible differences in visualized \ac{PDE} solutions, as shown in Figures \ref{fig:2d-single-deformation-results} and \ref{fig:3d-single-deformation-results}, and the animated versions provided in the Supplementary material.

A key strength of LPM is its rigorous mathematical framework, which has high potential for application to other \acp{PDE}, and can provide insights for physics-informed machine learning practitioners working with limited data and geometric variability. This is especially valuable given the growing body of work incorporating geometric variability into physics-informed learning, through shape descriptors as inputs \cite{regazzoni2022universal, costabal2024delta, sun2023physics, zhong2025physics}, or through specialized architectures \citep{gao2021phygeonet, dalton2023physics, peng2023physics, wurth2024physics, gao2022physics, kashefi2022physics, taufik2025latentpinns, karumuri2026physics, oldenburg2022geometry}. Building on these advances, LPM contributes an explicit mathematical formulation for incorporating shape gradients into the physics losses, thereby enabling more accurate gradient computations. Furthermore, we provided theoretical analysis of how shape boundary gradients are treated in conventional formulations, a perspective illustrated in Section \ref{subsec:shape-gradients} and Figure \ref{fig:2d-missing-shape-gradients}, offering complementary insights to guide future development in geometry-aware physics-informed learning.

The largest performance gains when applying latent \ac{PDE} mapping were observed for the rotational geometric families in 2D (\Grot, \GrotExt), and for the rotational families about the $y$- and $z$-axis (\Hroty, \HrotyExt, \Hrotz, \HrotzExt) in 3D. In these experiments, the LPM models yielded significantly lower mean relative $L_2$ errors compared to the ablation models. These results suggest that latent \ac{PDE} mapping could be beneficial in scenarios where geometric variability causes substantial boundary movements and changes in the \ac{PDE} solution. In contrast, for families where the boundary movement was more modest, such as for the expansion and shearing families in 2D (see Figure \ref{fig:2d-missing-shape-gradients}), the benefits of applying latent \ac{PDE} mapping were less obvious. Here, \Affine{} produced visibly accurate solutions that closely resembled the \ac{FEM} data (Figure \ref{fig:2d-single-deformation-results}). It should be noted that the \Affine{} network is architecturally similar to the PC-USM-Net formulation \cite{regazzoni2022universal} and the \ac{PINN} proposed by \citet{sun2023physics}, where global shape parameters are incorporated as additional inputs to represent geometric variations across the physical geometries. The comparable performance of \Affine{} and \LPM{} in these setting suggests that, for relatively simple and structured geometric variations,  network architectures with simple geometrical parameter inputs can effectively capture geometric variation and give accurate predictions on novel geometries, even without LPM.

The extension to rotational families in three dimensions provides further insights on the conditions under which latent \ac{PDE} mapping is advantageous. For \Hrotx{} and \HrotxExt, \LG{} yielded visibly accurate predictions that were comparable to the \LPM{} and \LPMDeepO{} (Figure \ref{fig:3d-single-deformation-results}). This observation suggests that mapping data to a predefined latent geometry without LPM \cite{li2023fourier, li2023geometry, yin2024scalable, zhong2025physics, mezzadri2023framework, regazzoni2022universal, burbulla2023physics} may be sufficient. However, it should be noted that in all of our experiments the myocardial fibers were aligned with the $x$-axis. Thus, rotations about the $x$-axis left the fiber orientation unchanged, resulting in axis-symmetric solutions with minimal variance. In contrast, rotations about the $y$- and $z$-axes (\Hroty, \HrotyExt{}, \Hrotz, \HrotzExt{}) altered the fiber orientation, producing more substantial changes to the Aliev-Panfilov solutions. For these rotation families, LPM yielded significant error improvements (Table \ref{tab:3d-single-deformations}). This suggests that the primary advantage of LPM lies in settings where geometric variations induce substantial changes to \ac{PDE} solutions.

Moreover, our results indicate that latent \ac{PDE} mapping remains effective when the geometries are described through a low-dimensional parameterization based on \ac{PCA} components (bottom panels of Table \ref{tab:2d-single-deformations} and \ref{tab:3d-single-deformations}). This is a particularly important result from an application perspective, as explicit deformation parameters are rarely available in practice. Instead, geometric variability is often characterized using low-dimensional shape descriptors, with \ac{PCA} being one of the most widely used approaches across a broad range of applications \cite{jolliffe2016principal, ovsjanikov2011exploration}, including cardiac modeling \cite{mauger2019right}. For example, \citet{yin2024scalable} employed \ac{PCA} scores as geometric descriptors in an operator-learning framework for predicting transmembrane activation times across populations of left ventricular geometries. Since latent \ac{PDE} mapping is formulated independently of the underlying neural network architecture, it could be integrated with such approaches to enforce a consistent physics-informed loss across geometries. This may improve data efficiency and enhance generalization to unseen anatomical variations.

Our quantification of the missing boundary shape-gradient contribution when latent \ac{PDE} mapping was not applied (Figure \ref{fig:2d-missing-shape-gradients}) provided additional insight into the magnitude of these neglected terms. In all 2D geometry families, the magnitude of the omitted boundary contribution (blue) exceeded the remaining physics gradients (orange), suggesting that conventional physics-informed formulations may disregard information that is relevant for learning across varying geometries. In Figure~\ref{fig:2d-missing-shape-gradients}, \Gexp{} and \Gshear{} exhibited smaller magnitudes in the missing boundary shape-gradient than for \Gnonlin{} and \Grot. Hence, the second term in \eqref{eq:reynold} was smaller for \Gexp{} and \Gshear{} than for \Gnonlin{} and \Grot. Looking at \eqref{eq:reynold}, the differences in magnitude could either be caused by differences in $\mathcal{R}(\vx, t, u)^2$ or in $\partial \vx / \partial s  \cdot \vn$. Moreover, the standard deviations in \Gnonlin{} were more pronounced than in the other families (\Gexp, \Gshear, \Grot). Since the same geometries in the given family were used for evaluating the shape-gradient at every epoch in Figure \ref{fig:2d-missing-shape-gradients}, the $\partial \vx / \partial s  \cdot \vn$ term should be constant across epochs. This suggests that $\mathcal{R}(\vx, t, u)^2$ varied more between epochs for \Gnonlin, thereby influencing the missing shape-gradient term more greatly than in the other families. For \Gexp, \Gshear, and \Grot, the standard deviations were smaller, indicating that $\mathcal{R}(\vx, t, u)^2$ varied less and that the quantification of the missing shape-gradient in these families were mainly caused by changes in the boundary motions $\partial \vx / \partial s  \cdot \vn$. While these results highlighted the potential importance of shape-gradient information, the precise mechanisms by which neural networks utilize these gradients during training remain unclear. Thus, there is a need for a deeper theoretical and empirical investigation of the role of shape gradients in future work to better understand their potential importance. 

Furthermore, our results show that latent \ac{PDE} mapping introduces some additional computational cost during training (Figure \ref{fig:computational-times}). However, this additional cost is modest and confined to the training phase, where the governing physics must be pulled back to the latent geometry. During inference, no such operations are required, leading to prediction times that are comparable to those of the ablation models. Moreover, the average cost of computing $\tF$ during mesh generation was $22.5 \pm 10.6$ and $318.4 \pm 36.3$ seconds per geometry in 2D and 3D, respectively. Although this preprocessing cost is expected to increase for more complex three-dimensional geometries, it remains a one-time expense that is incurred only for the training geometries. It should also be noted that the reported training and inference times (Figure \ref{fig:computational-times}) were measured using a single NVIDIA HGX H200 GPU on a shared compute cluster. Consequently, the reported runtimes should be interpreted as indicative rather than absolute, as modest variations may arise across different hardware configurations and system loads.

In our current work, we focused on two representative physics-informed learning frameworks, namely \acp{PINN} and \acp{PI-DON}. Although it may be tempting to compare \acp{PINN} and \acp{PI-DON} directly in Table \ref{tab:2d-single-deformations} and \ref{tab:3d-single-deformations}, the goal of this study is not to contrast different physics-informed frameworks, but rather to assess the performance gains obtained by applying latent \ac{PDE} mapping. Both \acp{PINN} and \acp{PI-DON} have distinct strengths and weaknesses and operate most effectively in different regimes. \Acp{PINN} have been shown to be particularly effective in sparse-data settings, such as inverse problems, where a full-field solution must be reconstructed from limited observations \cite{karniadakis2021physics, cuomo2022scientific}. \Acp{PI-DON}, on the other hand, are more suitable for learning across variable geometries, but require more solutions of the governing \ac{PDE} during training to accurately learn the governing operator \cite{wang2021learning}. Thus, although training a \ac{PINN} versus a \ac{PI-DON} differs fundamentally in terms of the learned representation, our results show that applying latent \ac{PDE} mapping enhances performance in both frameworks.

\subsection{Limitations and future work}
While latent \ac{PDE} mapping shows promising results for training physics-informed frameworks across various geometries with limited data, several limitations highlight opportunities for future research. A key limitation in our current work is that we assumed knowledge of the exact deformation gradient between the latent geometry and physical geometries. In many practical applications, the deformation gradient would need to be approximated numerically from geometric data, with the quality of the approximation potentially impacting the efficacy of applying LPM. Thus, future work can focus on extending latent \ac{PDE} mapping to operate in such settings, taking an important step toward real-world applicability.

Although our results demonstrate that latent \ac{PDE} mapping remains effective when geometries are represented through \ac{PCA}-based descriptors, further validation on more realistic and complex geometric variations is needed. One possible direction is to generate geometric populations using statistical shape models, which are commonly used in e.g., cardiac modeling to characterize anatomical variability in a population \cite{nagel2021bi}. Such datasets would provide a more realistic test of the method's ability to generalize across patient-specific anatomies. Alternatively, more flexible deformation models, such as B-spline-based parameterizations \cite{hasan2024b, mezzadri2023framework}, could be employed to generate geometries exhibiting localized and highly nonlinear shape variations. Furthermore, latent \ac{PDE} mapping could be combined with the $\Delta$-PINN \cite{costabal2024delta} to represent more complex geometries and manifolds. Evaluating latent \ac{PDE} mapping in these settings could provide further insight into its robustness and applicability to complex real-world problems.

Another limitation of the present study is that we have not systematically investigated the influence of the latent geometry on model performance. Although the proposed framework demonstrated robust performance across the considered test cases, the sensitivity of training and prediction accuracy to the choice of latent geometry remains unexplored and out of scope for the current work. Nevertheless, future work should evaluate alternative latent geometries and how the choice of latent geometry influences model performance. Moreover, we did not investigate the effect of finite element discretization and hence \ac{FEM} solution accuracy in our experiments. However, our discretization level was based on a previous benchmark study \cite{niederer2011verification} which reported only minor changes to \ac{FEM} solutions under mesh and time refinements beyond the level of our experiments. Also, any inaccuracy in \ac{FEM} solutions would have presumably affected both the LPM and ablation model results, and hence would have been unlikely to affect our study conclusions.

It should also be noted that the shape-gradients considered in this work were approximations that captured the sensitivity of the solution with respect to geometric variations and not the sensitivity of the network to a particular choice of a deformation map (see Appendix \ref{app:computational-graph}). While this formulation was sufficient for learning solution operators across varying geometries in our simplified experiments, real-world applications with more complex geometries might benefit from the explicit inclusion of the missing terms in \eqref{eq:computational-graph} involving the deformation gradient $\tF$ and Jacobian $J$. It is possible that the inclusion of the additional terms could have further improved the performance of the LPM models in our experiments. 

While latent \ac{PDE} mapping showed promising results for facilitating physics-informed learning across varying geometries in data-limited settings, this study did not investigate whether the same benefits persist when larger training datasets are available. Furthermore, the results suggest that LPM is most effective when geometric variations induce substantial changes in the underlying \ac{PDE} solutions, whereas its advantage is less pronounced for simpler geometric variations. Consequently, an important direction for future work is to establish a better understanding of the conditions under which LPM provides a significant benefit and to identify scenarios where its additional complexity may not be justified. Moreover, the reported experiments were conducted without a systematic evaluation across multiple random seeds. As a result, the influence of stochastic factors during training on the reported performance remains unquantified and should be assessed in future work to better characterize the robustness of the proposed approach. 

A final limitation of the present work is that the methodology has only been evaluated on a single governing equation, namely the Aliev-Panfilov \ac{PDE}. Although this \ac{PDE} provides a challenging benchmark due to its nonlinear, time-dependent dynamics and the presence of sharp gradients, the extent to which the observed benefits of latent \ac{PDE} mapping generalize to other classes of \acp{PDE} remains an open question. Additionally, even though our proposed framework naturally extends to time-dependent deformations, our current work solely consider static deformations. Thus, future work should therefore investigate the approach across a broader range of problems, as well as time-dependent geometric variations.

\section{Conclusion}
\label{sec:conclusion}
In this work, we presented latent \ac{PDE} mapping, a framework that pulls back geometry-specific \ac{PDE} and \ac{BC} residuals to a predefined latent geometry through the deformation gradient, thereby providing a consistent representation of both the governing physics and the solution data across varying geometries. By transforming a geometry-dependent learning problem into a parametric one, the proposed approach enables more accurate shape-gradient information and improves generalization to geometries outside the training distribution. Across both \acp{PINN} and \acp{PI-DON}, latent \ac{PDE} mapping consistently improved predictive performance, with the largest gains observed in settings where geometric variations induced substantial changes in the \ac{PDE} solutions.

Overall, the results demonstrate that latent \ac{PDE} mapping provides an effective mechanism for extending physics-informed learning frameworks with soft constraints to diffeomorphic varying geometries in both two and three dimensions. The framework is both \ac{PDE} and network architecture-agnostic and therefore has the potential to improve geometric generalizability for a broad class of physics-informed formulations.

\clearpage

\appendix

\section{Derivation of latent PDE mapping applied to the Aliev-Panfilov PDE}
\label{appendix:mapped_ap}

In the following section, we give a detailed derivation of how the Aliev-Panfilov \ac{PDE} in \eqref{eq:ap} is mapped from a physical geometry $\Omega(s)$ to a latent geometry $\Omega_0$. For convenience, we restate the equations over $\Omega(s)$ here as
\begin{equation}
    \begin{cases}
        \frac{\partial V}{\partial \tau} = \nabla \cdot (\tD \nabla V) - kV(V - a)(V - 1) - VW & \text{in } \Omega(s),  \\
        \frac{\partial W}{\partial \tau} = \left( \epsilon_0 + \frac{\mu_1 W}{V + \mu_2} \right) \left( -W-kV \left( V-a-1 \right) \right) & \text{in } \Omega(s), \\
        \tD  \nabla V \cdot \vn = 0 & \text{on } \partial \Omega(s). \\
    \end{cases}
    \label{eq:ap-appendix}
\end{equation}
The mapping is achieved by applying the deformation gradient $\tF(\vX, s)$ and the deformation Jacobian $J(\vX, s)$ to quantities in \eqref{eq:ap-appendix}, as well as performing a variable substitution $\vx \rightarrow \vX$ where $\vx \in \Omega(s)$ and $\vX \in \Omega_0$. The deformation gradient $\tF$ is given as
\begin{equation}
    \tF(\vX, s) = \nabla_{\vX}\Phi^{-1}(\vX, s)
\end{equation}
where $\Phi(\vX, s)$ is the continuous map between $\Omega(s)$ and the predefined latent geometry $\Omega_0$, while the deformation Jacobian is given as $J(\vX, s) = \text{det}(\tF)$.

We start by introducing how quantities in \eqref{eq:ap-appendix} are mapped when applying principles from nonlinear solid mechanics \citep{HolzaphelSolidMechanicsBook}. Quantities that do not involve any divergences or gradients are mapped directly through a volume change defined as
\begin{equation}
    \text{d}v = J \text{d}V
    \label{eq:volume-change-appendix}
\end{equation}
where $dv$ and $dV$ are infinitesimally small volume elements in $\Omega(s)$ and $\Omega_0$, respectively. Gradients of a scalar field $\phi$ are mapped as
\begin{equation}
    \nabla \phi(\vx,t) = \tF^{-T} \nabla \phi(\vX,t)
    \label{eq:scalar-gradient-relation-appendix}
\end{equation}
which is obtained by applying the chain rule to $\nabla \phi(\vx,t)$. Finally, Nanson's formula is used to map vector elements from $\Omega(s)$ to $\Omega_0$. The formula yields
\begin{equation}
    \text{d}s \vn = J \tF^{-T} \text{d}S \vN
    \label{eq:nanson-appendix}
\end{equation}
where $\text{d}s \vn$ and $\text{d}S \vN$ give the vector elements of infinitesimally small surface areas defined on $\Omega(s)$ and $\Omega_0$.

Next, we rewrite the divergence term in \eqref{eq:ap-appendix} in integral form as
\begin{equation*}
     \int_{\Omega(s)} \nabla \cdot (\tD \nabla V) \text{d}\Omega.
\end{equation*}
By applying Gauss's divergence theorem, we have that
\begin{equation}
    \int_{\Omega(s)} \nabla \cdot (\tD \nabla V) \text{d}\Omega = \int_{\partial \Omega(s)} \tD \nabla V \cdot \text{d}s \vn
    \label{eq:first-divergence-appendix}
\end{equation}
where $\partial \Omega(s)$ is the surface of $\Omega(s)$ and $\vn$ is the vector normal to the surface. We utilize the relationship of gradients in \eqref{eq:scalar-gradient-relation-appendix} and Nanson's formula in \eqref{eq:nanson-appendix}, such that the divergence term in \eqref{eq:first-divergence-appendix} can be expressed over $\Omega_0$ as
\begin{align}
    \int_{\partial \Omega(s)} \tD \nabla V \cdot \text{d}s \vn &= \int_{\partial \Omega_0} \tD \tF^{-T} \nabla V \cdot J \tF^{-T} \text{d}S \vN 
    \label{eq:first-mapped-appendix}
\end{align}
In the 2D case, we have that
\begin{equation*}
    \tD \in \mathbb{R}^{2x2}, \hspace{1cm} \tF \in \mathbb{R}^{2x2}, \hspace{1cm} \nabla V \in \mathbb{R}^{2x1}.
\end{equation*}
Hence, by assuming that $\tF$ is invertible, the terms in \eqref{eq:first-mapped-appendix} can be reorganized as
\begin{equation}
    \int_{\partial \Omega(s)} \tD \nabla V \cdot \text{d}s \vn = \int_{\partial \Omega_0} J \tF^{-1} \tD \tF^{-T} \nabla V \cdot \text{d}S \vN
\end{equation}
Finally, by applying Gauss's divergence theorem again, the divergence term in $\Omega(s)$ and $\Omega_0$ can be expressed as
\begin{equation}
    \int_{\Omega(s)} \nabla \cdot (\tD \nabla V) \text{d}\Omega = \int_{\Omega_0} \nabla \cdot \left( J \tF^{-1} \tD \tF^{-T} \nabla V \right) \text{d}\Omega_0
    \label{eq:mapped-lhs-appendix}
\end{equation}

By following the same procedure, the boundary condition in \eqref{eq:ap-appendix} can be rewritten as
\begin{equation}
    \int_{\Omega(s)} ( \tD \nabla V \cdot \vn)\text{d}\Omega = \int_{\Omega_0} (J \tF^{-1} \tD \tF^{-T} \nabla V \cdot \vN)\text{d}\Omega_0 \label{eq:ap-bc}
\end{equation}

The remaining parts of \eqref{eq:ap-appendix} do not include any divergences or gradients, and are mapped directly through a volume change as defined in \eqref{eq:volume-change-appendix}. Consequently, \eqref{eq:ap-appendix} can be expressed over $\Omega_0$ as
\begin{equation}
    \begin{cases}
        \frac{\partial}{\partial \tau} (JV) = \nabla \cdot (J \tF^{-1} \tD \tF^{-T} \nabla V) - J \left( kV(V - a)(V - 1) - VW \right) & \text{in } \Omega_0,  \\
        \frac{\partial}{\partial \tau} (JW) = J\left( \epsilon_0 + \frac{\mu_1 W}{V + \mu_2} \right) \left( -W-kV \left( V-a-1 \right) \right) & \text{in } \Omega_0, \\
        J \tF^{-1} \tD \tF^{-T} \nabla V \cdot \vN = 0 & \text{on } \partial \Omega_0, \\
    \end{cases}
\end{equation}

For time-independent mappings, we finally arrive at
\begin{equation}
    \begin{cases}
        \frac{\partial V}{\partial \tau} = \frac{1}{J} \nabla \cdot (J \tF^{-1} \tD \tF^{-T} \nabla V) - kV(V - a)(V - 1) - VW & \text{in } \Omega_0,  \\
        \frac{\partial W}{\partial \tau} = \left( \epsilon_0 + \frac{\mu_1 W}{V + \mu_2} \right) \left( -W-kV \left( V-a-1 \right) \right) & \text{in } \Omega_0, \\
        J \tF^{-1} \tD \tF^{-T} \nabla V \cdot \vN = 0 & \text{on } \partial \Omega_0.
        \label{eq:mapped-ap-appendix}
    \end{cases}
\end{equation}

\subsection{Computational graph used for the shape-gradient}
\label{app:computational-graph}
The physics and boundary condition loss, as defined by the mapped Aliev-Panfilov \ac{PDE} in \eqref{eq:mapped-ap-appendix}, are given as
\begin{align*}
    \mathcal{L}_{phys}^{LPM} &= \frac{1}{\mathcal{N}_{phys}} \sum^{\mathcal{N}_{phys}} [ \mathcal{R}^{LPM} (\vX, \tau, \hat{V}, \hat{W}, \tF, J ) ]^2,\\
    \mathcal{L}_{bc}^{LPM} &= \frac{1}{\mathcal{N}_{bc}} \sum^{\mathcal{N}_{bc}} [ \mathcal{R}^{LPM}(\vX, \tau, \hat{V}, \tF, J) ]^2
\end{align*}
where $\hat{V}, \hat{W}, \tF, J$ depend on $s$. During training, $\frac{\partial \mathcal{L}_{phys}^{LPM}}{\partial \theta}$ and $\frac{\partial \mathcal{L}_{bc}^{LPM}}{\partial \theta}$ are computed in order to update the network parameters $\theta$. In the following, we focus on $\frac{\partial \mathcal{L}_{phys}^{LPM}}{\partial \theta}$ and apply the chain rule to derive the computational graph, yielding
\begin{align*}
    \frac{\partial \mathcal{L}_{phys}^{LPM}}{\partial \theta}
    &= \frac{\partial \mathcal{L}_{phys}^{LPM}}{\partial \mathcal{R}} \left( \frac{\partial \mathcal{R}}{\partial \vX} \frac{\partial \vX}{\partial \theta} + \frac{\partial \mathcal{R}}{\partial \tau} \frac{\partial \tau}{\partial \theta} + \frac{\partial \mathcal{R}}{\partial \hat{V}} \frac{\partial \hat{V}}{\partial s} \frac{\partial s}{\partial \theta} + \frac{\partial \mathcal{R}}{\partial \hat{W}} \frac{\partial \hat{W}}{\partial s} \frac{\partial s}{\partial \theta} + \frac{\partial \mathcal{R}}{\partial \tF} \frac{\partial \tF}{\partial s} \frac{\partial s}{\partial \theta} + \frac{\partial \mathcal{R}}{\partial J} \frac{\partial J}{\partial s} \frac{\partial s}{\partial \theta} \right).
\end{align*}
Thus, the shape gradient $\frac{\partial \mathcal{L}_{phys}^{LPM}}{\partial s}$ is given as
\begin{equation}
    \frac{\partial \mathcal{L}_{phys}^{LPM}}{\partial s} = \underbrace{\frac{\partial \mathcal{L}_{phys}^{LPM}}{\partial \mathcal{R}} \frac{\partial \mathcal{R}}{\partial \hat{V}} \frac{\partial \hat{V}}{\partial s} + \frac{\partial \mathcal{L}_{phys}^{LPM}}{\partial \mathcal{R}} \frac{\partial \mathcal{R}}{\partial \hat{W}} \frac{\partial \hat{W}}{\partial s}}_{\text{solution sensitivity}} + \underbrace{\frac{\partial \mathcal{L}_{phys}^{LPM}}{\partial \mathcal{R}} \frac{\partial \mathcal{R}}{\partial \tF} \frac{\partial \tF}{\partial s} + \frac{\partial \mathcal{L}_{phys}^{LPM}}{\partial \mathcal{R}} \frac{\partial \mathcal{R}}{\partial J} \frac{\partial J}{\partial s}}_{\text{deformation map sensitivity}}
    \label{eq:computational-graph}
\end{equation}
where the first two terms describe how the solution $(\hat{V}, \hat{W})$ changes with the geometry, while the last two terms describe the sensitivity of the residual with respect to the choice of deformation map. Note that it is possible for the geometry $\Omega(s)$ to be realized through multiple equivalent deformation maps. In this work, we determined the deformation gradient $\tF$ and Jacobian $J$ apriori, precomputed them for each geometry and treated them as constant coefficient fields during training. Thus, $\tF$ and $J$ were not part of the computational graph during automatic differentiation, and the shape gradient considered in this work was approximated by
\begin{equation}
    \frac{\partial \mathcal{L}_{phys}^{LPM}}{\partial s} \approx \frac{\partial \mathcal{L}_{phys}^{LPM}}{\partial \mathcal{R}} \frac{\partial \mathcal{R}}{\partial \hat{V}} \frac{\partial \hat{V}}{\partial s} + \frac{\partial \mathcal{L}_{phys}^{LPM}}{\partial \mathcal{R}} \frac{\partial \mathcal{R}}{\partial \hat{W}} \frac{\partial \hat{W}}{\partial s}.
\end{equation}

\section{Synthetic data parameters}
\label{appendix:synthetic_data}
Table \ref{tab:opencarp_params} presents the parameter values used when generating synthetic data with the openCARP \ac{FEM} solver.

\begin{table}
    \caption{Parameter values used to create synthetic data. \Ac{PDE} parameters were selected in accordance with \citet{aliev1996simple} while conductivities were selected as proposed by \citet{niederer2011verification}.}
    \label{tab:opencarp_params}
    \begin{center}
        \begin{small}
                \begin{tabular}{lccccc}
                \toprule
                \multicolumn{1}{c}{Parameter}  &\multicolumn{1}{c}{Description} &\multicolumn{1}{c}{Value} \\
                \midrule
                $C_m$ &Membrane capacitance &\SI{1}{\micro F \cm^{-2}} \\
                $\beta$ & Surface area to volume ratio & \SI{0.14}{\meter^{-1}} \\
                $f_x, f_y, f_z$ & Fiber orientation & (1, 0, 0) $\in \Omega_0$ \\
                $\Delta t$ & Time resolution & \SI{0.005}{\ms} \\
                $I_{app}$ & Applied stimuli & \SI{5000}{\mu A \cm^{-2}} for \SI{0.2}{ms} \\
                \midrule
                $\sigma_{il}$ & Intracellular longitudinal conductivity & \SI{0.17}{Sm^{-1}} \\
                $\sigma_{it}$ & Intracellular transversal conductivity & \SI{0.019}{Sm^{-1}} \\
                $\sigma_{in}$ &Intracellular sheet-normal conductivity & \SI{0.019}{Sm^{-1}} \\
                $\sigma_{el}$ & Extracellular longitudinal conductivity & \SI{0.62}{Sm^{-1}} \\
                $\sigma_{et}$ & Extracellular transversal conductivity & \SI{0.24}{Sm^{-1}} \\
                $\sigma_{en}$ &Extracellular sheet-normal conductivity & \SI{0.24}{Sm^{-1}} \\
                \midrule
                $k$ & \ac{PDE} parameter & 8.0 \\
                $a$ & \ac{PDE} parameter & 0.15 \\
                $\varepsilon_0$ & \ac{PDE} parameter & 0.002 \\
                $\mu_1$ & \ac{PDE} parameter & 0.2 \\
                $\mu_2$ & \ac{PDE} parameter & 0.3 \\
                \bottomrule
                \end{tabular}
        \end{small}
    \end{center}
  \vskip -0.1in
\end{table}

\section{Hyperparameters and implementation details}
\label{appendix:model_configs}
All neural networks were implemented with \textit{PyTorch}, and experiments were run on NVIDIA HGX H200 GPUs. Table \ref{tab:models-config-pinns} and \ref{tab:models-config-deeponet} present the hyperparameters used in the \acp{PINN} and \acp{PI-DON}, respectively. Additionally, the number of \ac{FEM} space-time discretization points used to compute losses during training are given in Table \ref{tab:models-config-colloc}. Note that the \acp{PI-DON} included $N_{\mathrm{s}} = 14$ and $N_{\mathrm{s}} = 42$ sensor locations that provided initial transmembrane values $V_0$ to the branch network in 2D and 3D, respectively. 

\begin{table}[t]
    \caption{Overview of \ac{PINN} hyperparameters. Values in parentheses apply when PCA was used as a geometric descriptor.}
    \label{tab:models-config-pinns}
    \centering
    \begin{tabular}{llcccc}
        \toprule
        \multicolumn{2}{l}{}
            & \textsc{\LPM} & \textsc{\LG} & \textsc{\Affine} & \textsc{\PINN} \\
        \midrule
        \multicolumn{6}{l}{\textit{Architecture}} \\
        \quad Input dim      & 2D & 9\,(5) & 9\,(5) & 9\,(5) & 3 \\
                             & 3D & 13\,(6) & 13\,(6) & 13\,(6) & 4 \\
        \quad Hidden layers  & & \multicolumn{4}{c}{8} \\
        \quad Hidden dim     & & \multicolumn{4}{c}{64} \\
        \quad Output dim     & & \multicolumn{4}{c}{2} \\
        \quad Activation hidden layers & & \multicolumn{4}{c}{\texttt{tanh}} \\
        \quad Activation on $\hat{V}$   & & \multicolumn{4}{c}{\texttt{sigmoid}} \\
        \quad Activation on $\hat{W}$   & & \multicolumn{4}{c}{\texttt{softplus}} \\
        \midrule
        \multicolumn{6}{l}{\textit{Training}} \\
        \quad Epochs         & & \multicolumn{4}{c}{5000} \\
        \quad Batch size     & & \multicolumn{4}{c}{264} \\
        \quad Optimizer      & & \multicolumn{4}{c}{Adam} \\
        \quad Learning rate  & $\leq 100$ epochs & \multicolumn{4}{c}{$10^{-3}$} \\
                             & $> 100$ epochs    & \multicolumn{4}{c}{$10^{-4}$} \\
        \bottomrule
    \end{tabular}
\end{table}

\begin{table}[t]
    \caption{Overview of \ac{PI-DON} hyperparameters. Values in parentheses apply when \ac{PCA} was used as a geometric descriptor.}
    \label{tab:models-config-deeponet}
    \centering
    \begin{tabular}{llcc}
        \toprule
        \multicolumn{2}{l}{\textit{Hyperparameter}}
            & \textsc{\LPMDeepO} & \textsc{\LGDeepO} \\
        \midrule
        \multicolumn{4}{l}{\textit{Architecture}} \\
        \quad Branch input dim  & 2D & 20\,(16) & 20\,(16) \\
                                & 3D & 51\,(44) & 51\,(44) \\
        \quad Trunk input dim   & 2D & 3 & 3 \\
                                & 3D & 4 & 4 \\
        \quad Branch hidden layers  & & \multicolumn{2}{c}{6} \\
        \quad Trunk hidden layers   & & \multicolumn{2}{c}{6} \\
        \quad Branch hidden dim     & & \multicolumn{2}{c}{50} \\
        \quad Trunk hidden dim      & & \multicolumn{2}{c}{50} \\
        \quad Branch output dim     & & \multicolumn{2}{c}{50} \\
        \quad Trunk output dim      & & \multicolumn{2}{c}{100} \\
        \quad Output dim            & & \multicolumn{2}{c}{2} \\
        \quad Activation hidden layers        & & \multicolumn{2}{c}{\texttt{tanh}} \\
        \quad Activation on $\hat{V}$          & & \multicolumn{2}{c}{\texttt{sigmoid}} \\
        \quad Activation on $\hat{W}$          & & \multicolumn{2}{c}{\texttt{softplus}} \\
        \midrule
        \multicolumn{4}{l}{\textit{Training}} \\
        \quad Epochs            & & \multicolumn{2}{c}{5000} \\
        \quad Batch size        & 2D & \multicolumn{2}{c}{264} \\
                                & 3D & \multicolumn{2}{c}{128} \\
        \quad Optimizer         & & \multicolumn{2}{c}{Adam} \\
        \quad Learning rate     & $\leq 100$ epochs   & \multicolumn{2}{c}{$10^{-3}$} \\
                                & $> 2000$ epochs     & \multicolumn{2}{c}{$10^{-4}$} \\
        \bottomrule
    \end{tabular}
\end{table}

\begin{table}[t]
    \caption{\ac{FEM} space-time discretization points used during \ac{PINN} and \ac{PI-DON} training. The reduction in $\mathcal{N}_{data}$ for the 3D experiments relative to the 2D experiments is a consequence of the shorter simulation duration (500 ms compared with 620 ms). Points marked with $^\dagger$ were resampled at every epoch.}
    \label{tab:models-config-colloc}
    \centering
    \begin{threeparttable}
        \begin{tabular}{llcc}
            \toprule
            & & \textit{2D} & \textit{3D} \\
            \midrule
            $\mathcal{N}_{\mathrm{data}}$  & Measured data points         & 8610  & 6930\tnote{a} \\
            $\mathcal{N}_{\mathrm{phys}}$  & Colloction points$^\dagger$ & 430500 & 693000 \\
            $\mathcal{N}_{\mathrm{bc}}$    & Boundary points$^\dagger$    & 49200  & 79200  \\
            $\mathcal{N}_{\mathrm{ic}}$    & Initial points$^\dagger$     & 30  & 60   \\
            \bottomrule
        \end{tabular}
        \begin{tablenotes}
            \footnotesize
            \item[a] \acp{PI-DON} used $\mathcal{N}_{\mathrm{data}} = 20790$ in 3D.
        \end{tablenotes}
    \end{threeparttable}
\end{table}

\section{Numerical approximation of missing boundary gradients}
\label{appendix:numerical_analysis}

In the following section, we present details on how \eqref{eq:reynold} was discretized and numerically approximated. For convenience, we restate the equation here as 
\begin{equation}
    \frac{\partial \mathcal{L}_{phys}}{\partial s} = \int_{\Omega(s)} \frac{\partial}{\partial s}  \mathcal{R}(\vx, t, u, s)^2 \text{d}\Omega + \int_{\partial\Omega(s)} \mathcal{R}(\vx, t, u, s)^2 \frac{\partial \vx}{\partial s} \cdot \vn \text{d}S
\end{equation}
and define
\begin{align}
    I(s) &\equiv \int_{\Omega(s)} \frac{\partial}{\partial s} \mathcal{R}(\vx, t, u, s)^2 \text{d}\Omega \label{eq:I} \\
    B(s) &\equiv \int_{\partial \Omega(s)}  \mathcal{R}(\vx, t, u, s)^2 \frac{\partial \vx}{\partial s} \cdot \vn \text{d}S
    \label{eq:B}
\end{align}
such that
\begin{equation}
    \frac{\partial \mathcal{L}_{phys}}{\partial s} = I(s) + B(s).
\end{equation}

Next, we discretize the terms and make a numerical approximation using Monte Carlo for integrals and central finite differences for derivatives. We assume that the spatial positions are uniformly distributed and normalize with respect to the area/boundary, such that

\begin{equation}
        I(s_k) \approx \frac{1}{N_I(s)} \sum_i^{N_I(s)} \sum_j^{\mathcal{T}} \frac{\mathcal{R}(\vx_i, t_j, u_{ij}, s_k + \Delta s)^2 - \mathcal{R}(\vx_i, t_j, u_{ij}, s_k - \Delta s)^2}{2 \Delta s}
\end{equation}

and

\begin{equation}
    B(s_k) \approx \frac{1}{N_B(s)} \sum_i^{N_B(s)} \sum_j^{\mathcal{T}} \mathcal{R}(\vx_i, t_j, u_{ij}, s_k)^2 \underbrace{\frac{\vx_i(s_k + \Delta s) - \vx_i(s_k - \Delta s)}{2 \Delta s} \cdot \vn_i(s)}_{\text{boundary movement}}
    \label{eq:boundary-movement}
\end{equation}

where $s_k$ is the $k$-th value in a set of shape parameters given as $s = \{s_1, s_2, ... , s_K\}$. Moreover, $N_I(s)$ and $N_B(s)$ gives the number of spatial positions used to evaluate the two terms and $\mathcal{T}$ is the total number of time steps. Thus, our discretized version for the $k$-th shape value is given as

\begin{equation}
    \frac{\Delta \mathcal{L}_{phys}}{\Delta s_k} = I(s_k) + B(s_k)
    \label{eq:s_k}
\end{equation}

The magnitude of the overall change for the shape parameters ($\frac{\Delta \mathcal{L}_{phys}}{\Delta s}$) was computed by applying the $L_2$ norm to \eqref{eq:s_k}. Finally, we computed $\frac{\Delta \mathcal{L}_{phys}}{\Delta s}$ for each training geometry in a family, and represented the numerical approximation of $I$ and $B$ as the mean across the given training geometries. Here, $I$ represents the numerical computation of $\mathcal{L}_{phys}^{conv}$ and $B$ represents the missing boundary information when latent \ac{PDE} mapping is not applied (see Figure \ref{fig:2d-missing-shape-gradients}). We used $\Delta s = 10^{-6}$ in all computations.

\clearpage


\section*{Acknowledgments}
This work has received significant support from the Kristiania-HPC infrastructure, financially sponsored by Kristiania University of Applied Sciences, and the ProCardio center for Innovation funded by the Research Council of Norway, project 309762.

\section*{Data statement}
The simulated FEM electrophysiology data is publicly available at \href{https://doi.org/10.5281/zenodo.20928054}{https://doi.org/10.5281/zenodo.20928054}.

\bibliographystyle{cas-model2-names}


\end{document}